\documentclass[pdflatex,sn-mathphys-num]{sn-jnl}

\usepackage{amsmath, amsthm, amssymb, calrsfs, wasysym, verbatim, bbm, color, graphics, geometry}

\usepackage{array}
\usepackage{adjustbox}
\usepackage{amsfonts}
\usepackage{amsmath}
\usepackage{booktabs}
\usepackage{colortbl}
\usepackage{forest}
\usepackage{gensymb}
\usepackage{graphicx}
\usepackage{fontawesome}
\usepackage{multirow}
\usepackage{tabularx}
\usepackage{tikz} 
\usepackage{pgf, pgfsys, pgffor} 
\usepackage{pgfplotstable} 
\usepackage{pifont}
\usepackage{url}
\usepackage{wrapfig}
\usepackage{lineno}
\usepackage{pifont}  
\newcommand{\Letter}{\ding{41}}

\newif\ifshowpengfeicoms
\showpengfeicomsfalse
\newcommand{\pengfeicom}[1]{\ifshowpengfeicoms\textcolor{blue}{#1}\else#1\fi}

\newif\ifshowjiamingcoms
\showjiamingcomsfalse
\newcommand{\jiamingcom}[1]{\ifshowjiamingcoms\textcolor{green}{#1}\else#1\fi}

\newif\ifshowxintaocoms
\showxintaocomsfalse
\newcommand{\xintaocom}[1]{\ifshowxintaocoms\textcolor{red}{#1}\else#1\fi}

\newcommand{\cmark}{\textcolor{green}{\ding{51}}}%
\newcommand{\xmark}{\textcolor{red}{\ding{55}}}%

\newcommand{\eg}{\textit{e.g.}}
\newcommand{\ie}{\textit{i.e.}}
\newcommand{\gray}{\rowcolor[gray]{.95}} %

\newcommand{\lnk}[1]{\href{#1}{\faExternalLink}}

\newcolumntype{Y}{>{\centering\arraybackslash}X}
\newcolumntype{L}{>{\raggedright\arraybackslash}X}
\newcolumntype{s}{>{\raggedright\arraybackslash}m{0.09\hsize}}
\newcolumntype{a}{>{\raggedright\arraybackslash}m{0.1\hsize}}
\newcolumntype{b}{>{\raggedright\arraybackslash}m{0.08\hsize}}
\newcolumntype{d}{>{\raggedright\arraybackslash}m{0.06\hsize}}

\theoremstyle{thmstyleone}%
\theoremstyle{thmstyletwo}%

\theoremstyle{thmstylethree}%

\usepackage[dvipsnames]{xcolor}

\begin{document}


\title[Simulating the Visual World with Artificial Intelligence: \\
A Roadmap]{Simulating the Visual World with Artificial Intelligence:\\
A Roadmap}


\author{
\begin{center}
\fnm{Jingtong} \sur{Yue}\textsuperscript{1}\quad
\fnm{Ziqi} \sur{Huang}\textsuperscript{2†}\quad
\fnm{Zhaoxi} \sur{Chen}\textsuperscript{2}\quad
\fnm{Xintao} \sur{Wang}\textsuperscript{3}\quad\\
\fnm{Pengfei} \sur{Wan}\textsuperscript{3}\quad
\fnm{Ziwei} \sur{Liu}\textsuperscript{2\Letter}\\
\vspace{5pt}
$^{1}$Robotics Institute, Carnegie Mellon University\\
$^{2}$S-Lab, Nanyang Technological University\\
$^{3}$Kling Team, Kuaishou Technology
\end{center}

\color{Magenta}\url{https://world-model-roadmap.github.io/}

}





\abstract{The landscape of video generation is shifting, from a focus on generating visually appealing clips to building virtual environments that support interaction and maintain physical plausibility. These developments point towards the emergence of video foundation models that function not only as visual generators but also as implicit world models, models that simulate the physical dynamics,  agent-environment interactions, and task planning that govern real or imagined worlds.
    This survey provides a systematic overview of this evolution, conceptualizing modern video foundation models as the combination of two core components: an implicit world model and a video renderer. The world model encodes structured knowledge about the world, including physical laws, interaction dynamics, and agent behavior. It serves as a latent simulation engine that enables coherent visual reasoning, long-term temporal consistency, and goal-driven planning. The video renderer transforms this latent simulation into realistic visual observations, effectively producing videos as a ``window" into the simulated world. We trace the progression of video generation through four generations, in which the core capabilities advance step by step, ultimately culminating in a world model, built upon a video generation model, that embodies intrinsic physical plausibility, real-time multimodal interaction, and planning capabilities spanning multiple spatiotemporal scales.
    For each generation, we define its core characteristics, highlight representative works, and examine their application domains such as robotics, autonomous driving, and interactive gaming.
    Finally, we discuss open challenges and design principles for next-generation world models, including the role of agent intelligence in shaping and evaluating these systems. 
}




\maketitle
\makeatletter
\renewcommand\@makefntext[1]{\noindent #1}
\footnotetext{† project lead. \Letter~corresponding author.}
\makeatother

\section{Introduction}\label{sec:introduction}
\subsection{Motivation}
    \label{sec:intro_motivation}
World models, which aim to simulate the real world, have long posed significant challenges in artificial intelligence, influencing various applications, such as robotics, autonomous driving, and gaming. 
    While the specific capabilities required for world models are numerous and not yet precisely defined, approaches such as 3D generation~\cite{lumagenie2025, liu2025unleashing, huang2024placiddreamer, yang2019pointflow, cao2025physx}, 3D/4D scene generation~\cite{sun20253d, hsu2025programs, yang2024holodeck, yu2025wonderworld, yu2024wonderjourney, huang2025omnix, huang2025terra, zhaoxi20254dnex}, and video generation demonstrate one or more relevant abilities, such as motion dynamics, interaction and controllability, visual quality, 3D consistency, and generation efficiency. Consequently, these approaches have been adopted in recent works as potential pathways towards realizing world models.

    While these approaches each excel in partial capabilities towards world modeling, video generation offers one direct and comprehensive pathway, which may be a promising tool for building a world model.
    From a cognitive science perspective, vision is the dominant sensory modality through which both humans and embodied agents perceive, learn, and reason about the world. Visual streams not only convey spatial layout and object properties but also encode temporal dynamics and causal relationships crucial for prediction and planning. Even complex 3D or 4D simulations can be rendered into videos or images for interpretation, meaning that any human or embodied agent grounds its understanding in visual sequences. This intrinsic reliance on visual representation makes video generation a uniquely natural and information-rich foundation for constructing world models.

     Recent advances in end-to-end video generation indicate that such models can now serve as high-quality visual renderers, enabling the exploration of video-based approaches to world modeling. Recent advancements in techniques, such as diffusion models~\cite{peebles2023dit, podell2023sdxl, esser2024sd3, song2020denoising} and autoregressive transformers~\cite{parmar2018imagegeneration, razavi2019vqvae2}, have made it possible to generate high-quality videos grounded in fundamental world knowledge.
    Current methods~\cite{sora, teng2025magi, wan2025wan, peng2025opensora2, kong2024hunyuanvideo, xu2024easyanimate, bao2024vidu, yin2025causvid, polyak2024moviegen, huang2025selfforcing, klingai2024kling, yang2024cogvideox, lin2025apt2, agarwal2025cosmos, chen2025skyreels, gao2025seedance, mirage2025} based on these backbones are now capable of producing long-duration, high-quality videos with superficial faithfulness, endowing them with the ability to simulate real-world environments with high fidelity and incorporate multi-modal conditioning. As a result, there is an increasingly evident trend of utilizing video generation models as world models.

    \begin{figure}[!t]
      \centering
      \includegraphics[width=\linewidth]{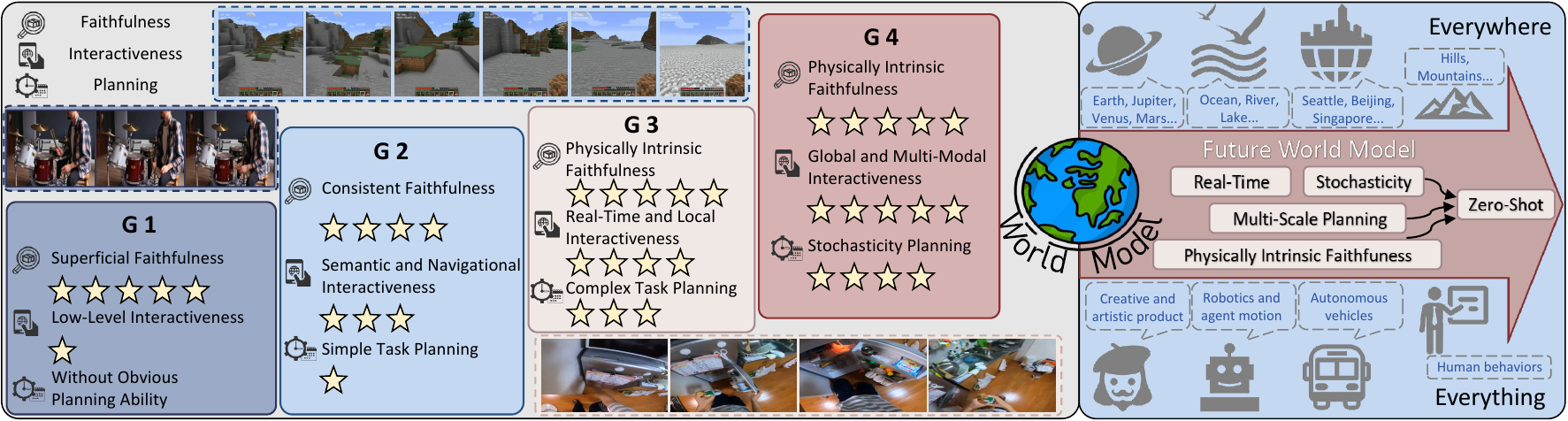}
      \caption{\textbf{Overview of 4 Generations and 3 Core Capabilities from Video Generation to World Model.} The figure illustrates the key capabilities emphasized in the first through third generations of world models, as well as our insight for future world models. We outline a long-term vision of world models that can simulate a broad range of environments across multiple spatial and temporal scales. The figure highlights four foundational characteristics: real-time responsiveness, stochasticity, multi-scale planning, and intrinsic physical faithfulness. These collectively support the long-term goal of zero-shot generalization.}
      \label{fig:cap}
    \end{figure}
    
    In this survey, we define a physical world model as a digital engine that encodes comprehensive world knowledge to simulate real‑world dynamics according to intrinsic physical laws. Such models serve both as high‑fidelity simulators for advancing domains like robotics, autonomous driving, and gaming, and as controlled testbeds for training and evaluating intelligent agents under realistic conditions. 

\begin{figure}[!t]
      \centering
      \includegraphics[width=\linewidth]{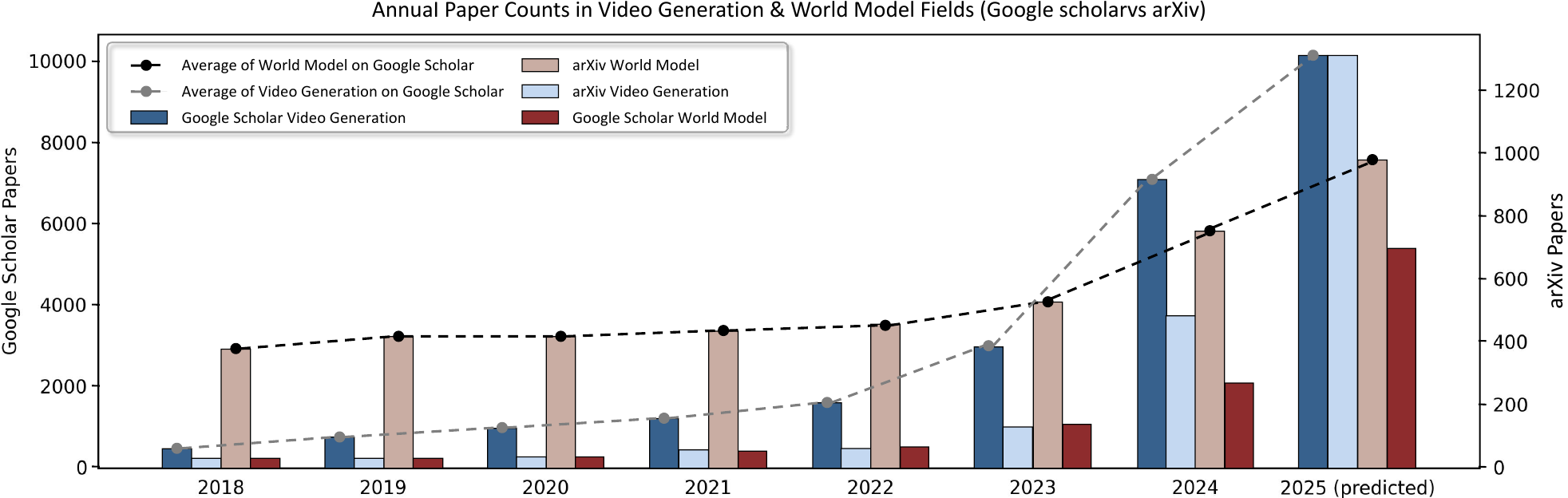}
      \caption{\textbf{Overview of Annual Papers and Articles Paper Counts in Video Generation \& World Model Fields.} The article count was derived from searches conducted using the fixed keyword combination “video generation” and “world model” from Google Scholar and arXiv. }
      \label{fig: paper_amount}
    \end{figure}

     Recent breakthroughs~\cite{assran2025vjepa2, mirage2025, klingai2024kling, sora, pika23, jimengai24jimeng, yang2024cogvideox, lin2025apt2, agarwal2025cosmos, alhaija2025cosmostransfer1, runway2025gen4, Veo325, chen2025skyreels, stepvideo25jieyue,gao2025seedance, bao2024vidu, luma24, yin2025causvid, hailuo0225minmax, teng2025magi, wan2025wan, kong2024hunyuanvideo, fan2025vchitect} in video generation models, driven by improvements in diffusion models~\cite{hatamizadeh2024diffit, zhang2023improved, song2023lossguided, mardani2023variational, song2023pseudoinverse, lim2023score, tashiro2021csdi, song2020ddim, zhu2025asdm, zhong2025vfrtok, shi2025latent, zhong2025decoupling}, autoregressive backbones~\cite{meng2020autoregressive}, variational autoencoders~\cite{zhao2019infovae, zhao2017learning}, image generation techniques~\cite{rombach2022high, dang2025personalized, zhang2023diffcollage, ye2023affordance, balaji2022ediff, meng2021improvedar, meng2021sdedit, luma24, zhao2017towards, liu2025rpe, shen2024sgadapter, zeng2024jedi, ho2020denoising, si2024freeu, huang2024reversion, shi2025svg, xian2025free, kumbong2025hmar, gordon2025stencil}, controllable image generation ~\cite{zhang2023controlnet,atzmon2024edifyimagediff, sinha2021d2c,mou2024t2i,huang2023collaborative, wang2023styleadapter}, and enhancements in training or inference efficiency~\cite{zhou2025inductive, song2025ideas, xu2024agg, ozturkler2023smrd, he2025scaling, wu2025vmoba,zhang2025Jenga, shi2025diffmoe, yang2025efficient, yin2025towards, han2024agent, ju2025fulldit, he2025fulldit2, zhang2025training, zhengyao2025dcm}, as well as more flexible condition injection modules~\cite{zhang2025motioncrafter, zhang2023controlnet}, mark a pivotal moment for the field. 
     To this end, recent progress further expands the capabilities of video generation across video reasoning and alignment~\cite{huang2025vchain, liu2025improving, li2025gran, wang2025monet, yin2025sea, cheng2025realdpo, yuan2024instructvideo}, long-horizon storytelling and film generation~\cite{wang2025mavis, he2024storytelling, huang2025filmaster, zhuang2024vlogger, zhao2024moviedreamer, xie2024dreamfactory,wu2025automated, he2025cut2next}, high-resolution video generation~\cite{qiu2025cinescale, he2024venhancer}, personalized or multi-concept video customization~\cite{huang2025conceptmaster, wu2025customcrafter}, and test-time scaling (TTS)~\cite{he2025scaling}. Collectively, these advances reinforce the emerging view that modern video generation systems are becoming fundamental components for constructing world models.
    Moreover, with the rapid advancement of technologies such as virtual reality (VR) and embodied AI, the integration of world models into interactive, real-time environments has become increasingly feasible. These technological trends, together with the maturation of video generation methods, indicate that we are on the cusp of a new era where world models will play a central role in shaping autonomous systems, intelligent agents, and immersive virtual environments. This trend is further evidenced in Figure~\ref{fig: paper_amount}, which illustrates the dynamics of research attention over recent years. Specifically, while discussions and mentions of world models have been present since 2018, they have remained relatively steady in terms of annual publication volume. While starting in 2024, video generation witnessed an explosive surge in both technical advances and the number of related works, which in turn catalyzed a renewed wave of progress in world models. This pattern highlights a clear interdependence: the rapid advances in video generation are not merely parallel developments but are becoming key enablers that support and accelerate the evolution of world models. Consequently, the present moment is particularly critical for the community to systematically discuss the evolution from video generation to world models, and to chart their future directions as an integrated research frontier.

    Despite these advances, challenges remain at both the conceptual and structural levels. Conceptually, the definition of a “world model” remains ambiguous, making it difficult to unify perspectives and evaluate progress. Structurally, the field lacks a well-established taxonomy to organize modeling capabilities, developmental stages, and potential trajectories. These gaps highlight an urgent need for a systematic elucidation and comprehensive survey to consolidate existing knowledge and guide the next stage of research. 
    Current surveys and benchmarks~\cite{kong20253d, wang2025survey, xing2024survey, yu2025surveyigv, dal2024jointadsurvey, zhu2024issoraasurvey, fu2024exploringsurvey, melnik2024videosurvey, lin2025exploringsurvey, guan2024adinitialsurvey, liu2025generativesurvey, cho2024soraasansgisurvey, sun2024fromsorawecanseesurvey, yu2025position, duan2025worldscore, qin2024worldsimbench, li2025worldmodelbench,zheng2025vbench2, zhang2025evaluationagent} have laid important groundwork by summarizing related methodologies, datasets, and applications. However, there remains a need to explicitly clarify what aspects have been thoroughly addressed, and which areas, such as real-time integration, controllable video-to-world pipelines, or comprehensive evaluation metrics, are still underexplored. Accordingly, this survey aims to chart a clear path from video generation towards comprehensive world modeling, providing guidance for future research and development in this emerging field.

    \jiamingcom{In this paper, we begin with formal definitions and provide a detailed discussion centered around the taxonomy of world models.
    We define a physical world model as a digital simulation engine embedded with comprehensive world knowledge, capable of predicting the next scene conditioned on environmental states and contextual priors.
    Each prediction step can be represented as a triplet \textit{(Current Scene,Navigation Mode,Prior Information)}, which collectively determine the evolution of the simulated environment.
    The model focuses on capturing the causal and spatiotemporal dynamics of the physical environment, while external inputs such as navigation modes or actions act as stimuli that perturb or impact the environment’s evolution.
    Hence, the world model is formulated as an interactive environment system that responds to external interventions without explicitly modeling the decision-making process that generates them.
    Based on this definition, we systematically analyze how video generation models have evolved towards world modeling and propose a four-generation taxonomy according to model capability, as illustrated in Figure~\ref{fig:cap}}.

    \begin{itemize}
    \item Generation 1 - Faithfulness: Accurate Simulation of the Real World; 
    \item Generation 2 - Interactiveness: Controllability and Interactive Dynamics; 
    \item Generation 3 - Planning: Modeling the Future Evolution of Complex Systems;
    \item Generation 4 - Stochasticity: Modeling Outlier and Low-Probability Events.
    \end{itemize}
    
    We categorize and analyze existing approaches based on their foundation model ability, navigation mode types, application domains, and conditioning strategies. This perspective helps clarify how navigation signals influence video generation and what architectural designs are most effective for different world modeling tasks. Such a clear generational breakdown allows for a systematic evaluation of the progress in world modeling and facilitates a discussion of the remaining challenges and gaps between current video generation systems and ideal world models.

    \noindent
    We summarize our contributions as follows:
    \begin{itemize}
    \item A global taxonomy of world models: We propose a four-generation taxonomy for categorizing the evolution of video generation towards world models, grounded in the world model’s core capabilities, faithfulness, interactiveness, and planning.

    \item Clarification of the definition of world model:
    We define the core task as next-scene video prediction, and provide a formal equation that characterizes the world modeling process in terms of input, internal state, and output.

    \item Formal definition of navigation modes: We define the scope and characteristics of navigation modes, distinguishing them from spatial conditions to prevent conceptual overlap and to clarify their differences in interaction flexibility.
    
    \item Future perspectives: We discuss the key capabilities that must be achieved for video generation models to evolve into fully-fledged world models, offering insights for future research direction.
    \end{itemize}

    \subsection{Position}
    \label{sec:intro_position}

    \xintaocom{World models~\cite{lecun2022wm, ha2018worldmodel} have traditionally been regarded as tools enabling AI agents to perceive and interact with their environment, often inspired by human cognition and grounded in so‑called “common sense.” In this survey, we distinguish two complementary perspectives, a physical axis emphasizing external dynamics and a mental axis emphasizing internal simulation and intention modeling, following the conceptual distinction introduced by Huang in his blog post~\cite{huang2025towardsvwm}.}

    From this standpoint, the physical world model represents a more fundamental and global conceptualization, aiming to capture both the evolution and the intrinsic laws of the physical world. In contrast, the mental world model~\cite{kessler2023effectiveness, xu2022learning, lu2022challenges, ball2021augmented, kolev2024efficient} can be regarded as a specialized internal cognitive framework that may emerge in higher‑generation physical world models, serving to represent an agent’s internal states, intentions, and preferences. 

    \pengfeicom{A clear distinction between these two perspectives is essential, both philosophically and in terms of their representational forms and capability requirements. Philosophically, the distinction mirrors the classical debate between subjective idealism and mechanistic materialism. The physical world model aligns with the latter, seeking to explain the world through objective, immutable physical laws and focusing on external dynamics independent of the agent’s subjective experience. In contrast, the mental world model reflects a subjective, intentional stance, emphasizing the role of perception and intention in shaping the agent’s understanding of the world, consistent with subjective idealism, which holds that objective events may be influenced by an agent’s will or actions. As for capability demand, the physical world model emphasizes adherence to real‑world physical laws, interaction grounded in intrinsic world knowledge, and essential planning based on objective dynamics. The mental world model, on the other hand, should possess abilities such as semantic understanding, active interaction, and counterfactual reasoning, enabling it to think and act more like human beings. Because of these fundamental differences, the mental world model cannot be conflated with the physical world model in our definition; instead, we explicitly delineate the scope and capabilities of the physical world model.}
    \subsection{Scope}
    \label{sec:intro_scope}
    
   \xintaocom{This survey focuses on methods that advance video generation models towards world models. The objective of world model powered by video generation methods is to acquire structured world knowledge that can, in turn, enhance AI agents’ perception, reasoning, and planning capabilities. We consider world models as multimodal perceptual, interactive, and predictive systems that are capable of capturing the underlying dynamics, spatial structure, and semantics of the environment. These representations can support a broad spectrum of downstream tasks, including but not limited to visual planning, counterfactual reasoning, generalization to novel scenarios, and cross-modal understanding, enhancing the performance of agents and promoting research in physics-related fields. In addition, the broader usages of powerful world knowledge representation are still worth exploring.}

    \begin{figure}[!t]
      \centering
      \includegraphics[width=0.4\linewidth]{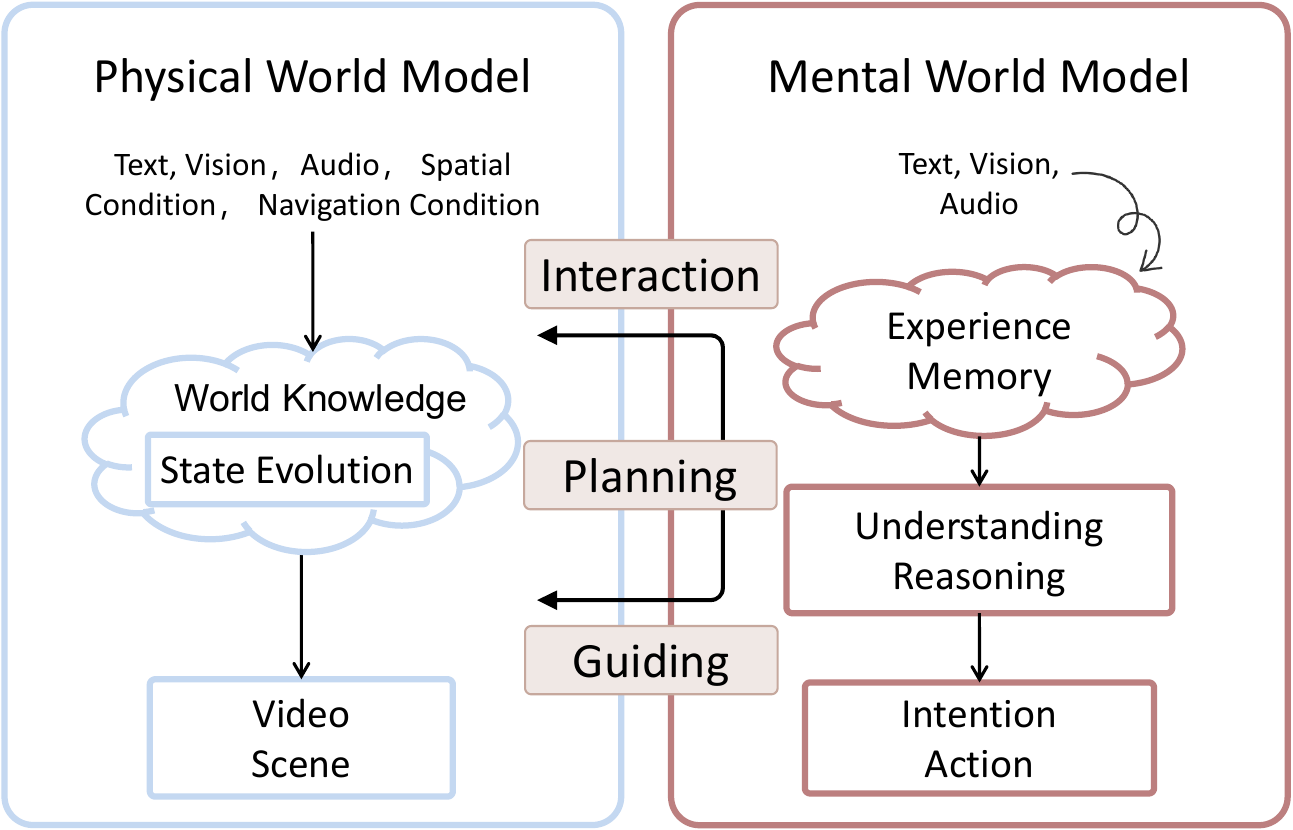}
      \caption{\textbf{The Characteristics of Physical World Model and Mental World Model.} \pengfeicom{This figure highlights the distinct inputs, internal processes, and outputs, as well as the interaction through perception, planning, and guidance between the physical world model and the mental world model.}}
      \label{fig: phy_men}
    \end{figure}

    \section{Problem Definition and Taxonomy}
    \label{sec:definition}
    
\begin{figure}[!t]
      \centering
      \includegraphics[width=\linewidth]{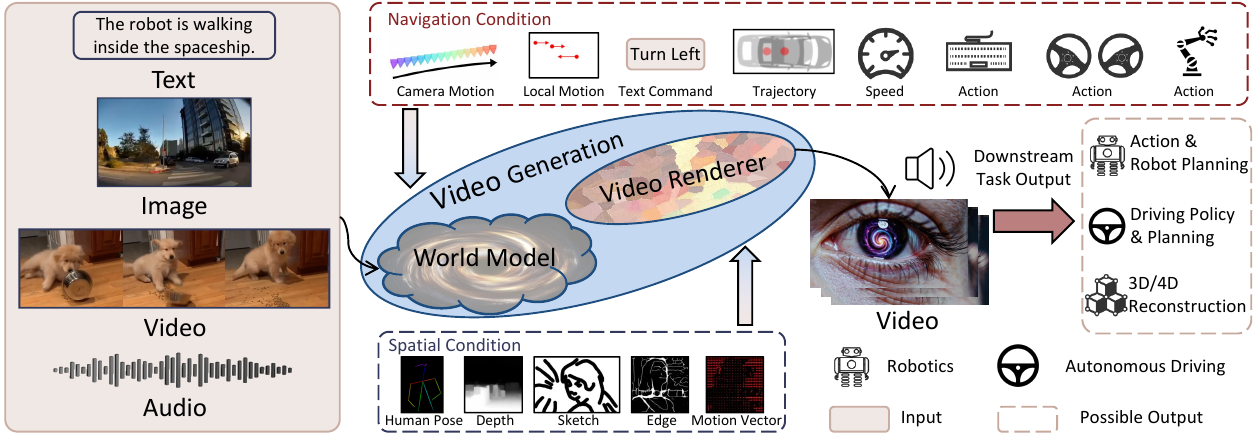}
      \caption{\pengfeicom{\jiamingcom{\textbf{Overview of the World Model Defined in this Paper.} The world model must take inputs such as text, images, videos, audios or their combinations. It may also incorporate external conditions for interaction, including spatial conditions and navigation conditions. A video generation model is leveraged to process the intermediate state representations to produce video outputs, while other task-specific outputs may also be generated depending on the downstream application.}}}

      \label{fig:definition}
    \end{figure}

    \xintaocom{\jiamingcom{Inspired by the concept of digital cousins from policy learning~\cite{dai2024automated} and sim-to-real research, where virtual environments are designed to share semantic and geometric affordances with the real world without being exact replicas. In our context, we define a physical world model as a digital cousin of the real world—one that embeds objective world knowledge and generates observable futures under external conditioning, as illustrated in Figure \ref{fig:definition}. Unlike a digital twin that replicates a specific world instance, a digital cousin emphasizes distributional realism, the capacity to simulate diverse, physically plausible yet semantically varied worlds, enabling world models to generalize beyond faithful reproduction. A world model not only simulates the causal dynamics and spatiotemporal evolution of the environment but also encodes the behaviors, interactions, and goals of agents within it. 
    In practice, the world model is initialized by external inputs, including textual prompts $T$ , current observations $O$ (images or video clips),  and other external interventions that stimulate or guide how the environment evolves, such as audio signals $Au$, navigation modes $N$ (actions, text commands, or trajectories), and spatial conditions $X$. These inputs define the starting state of the simulation and guide its evolution. Formally, the world mode is defined as follow:}}

    \begin{equation}
    V_{1:T} = \mathcal{G}(I), \quad I = \{T, O, Au, N, X\}
    \end{equation}
    \xintaocom{\jiamingcom{\pengfeicom{where $\mathcal{G}$ denotes the video generation model and $I$ represents the multimodal input space. The world model based on video generation model pipeline maps external inputs $I$ to a sequence of observable video frames $V_{1:T}$, which is a stochastic generative process.}}}

     \xintaocom{\pengfeicom{More specifically, under the decomposition, the world model corresponds to the latent representation $S_{t}$ together with the transition function $F$ (i.e., representation) which captures the model’s internalized world knowledge (dynamics, object affordances, agent intents, etc.) and compute $S_{t+1}=F(S_{t}, I_{t})$. The video renderer is the function $R$ that translates those internal world states into pixel-level or perceptual outputs through $V_{t+1} = R(S_{t+1})$.}} \xintaocom{\pengfeicom{Returning to the high-level view, although we conceptualize a video generation model as the combination of a world model and a video renderer, in practice the internal state of the world model is implicit. In other words, the video generation process still manifests as a single, monolithic mapping process  from input $I$ to output videos $V_{1:T}$.}}

     \xintaocom{\jiamingcom{Conceptually, the world model serves the same functional role as the environment dynamics in a Markov Decision Process (MDP). During training, exposed to large-scale multi-view and multi-temporal data, the model can approximate a fully observable environment where the latent world state $S_{t}$ is effectively known. The learned transition $F(S_{t}, I_{t})$ thus behaves analogously to an environment transition function $F(S_{t+1} | S_{t}, A_{t})$, where $A$ denotes the action space, in a fully observable MDP, forming an objective world prior. During inference, the model receives only partial conditioning signals (text, image, audio, action etc.), corresponding to partial observations of the true latent state. The resulting generative process therefore aligns with a Partially Observable MDP (POMDP), formulated as $F(S_{t+1}, O_{t+1} | S_{t}, A_{t})$. This dual interpretation reconciles the apparent contradiction between objectivity and subjectivity:
    training instills objective physical knowledge, while inference performs subjective reasoning grounded on that learned prior.
    It also clarifies that while the world model functions as an objective simulator of latent physics, its operation at inference time is conditioned by subjective, agent-like observations, bridging the two perspectives within a single unified framework.}}

    \subsection{\xintaocom{Taxonomy of Four Generations from Video Generation to World Model} }
    \label{sec:definition_taxonomy}

    In this section, we provide a brief version of the video generation to world models taxonomy, highlighting the development of core capabilities in each generation. For detailed comparisons, please refer to Section~\ref{sec: roadmap} and Figure~\ref{fig:generation_detail}. \xintaocom{The four generations form an evolutionary ladder, that each subsequent level extends the previous one along the same capability axes rather than representing independent types.}

    \begin{itemize}
    
    \item Generation 1 - Faithfulness: Superficial Simulation of the Real World
    
    In this generation, world models acquire the general ability of video generation models, enabling them to generate short videos lasting 2 to 5 seconds, barely satisfying human visual quality. The model also supports basic interactions conditioned on spatial signals with low flexibility and basic condition-video consistency. However, only basic commands and movements can be generated, and the ability to plan has yet to develop. These models typically use text, image, or video inputs as the initial state of the virtual world and spatial conditions as control signals.
    
    \item Generation 2 - Interactiveness: Controllable and Interactive Simulation of the Real World
    
    In this generation, world models build upon their superficial physical
    plausibility to acquire basic 2D and 3D reasoning capabilities, as well as more flexible and navigational controllability. They are now capable of generating long videos with richer content complexity, better visual realism, and perfect temporal and video-text consistency.  
    Moreover, simple task planning is developed to support basic task-oriented planning without physical evolution capability.
    These models are increasingly able to accept diverse conditioning and navigation mode inputs.

    \item Generation 3 - Planning: Real-Time and Complex Prediction of the Real World

    In this generation, planning has made significant advancements and emerged as a key capability of world models, that is, given an initial state, they can autonomously generate infinite or arbitrarily long videos for complex tasks based on intrinsic physical knowledge with real-time controllability. Models of this generation are expected to achieve real-time and subject-centric video generation with external control signals.

    \item Generation 4 - Stochasticity: Low-Probability and Multi-Scale Modeling of the Real World 
   
    In this generation, world models achieve a balance between modeling regular, rule-based environments and capturing low-probability events. They are now capable of representing rare occurrences, such as genetic mutations, unexpected accidents, financial crises, and volcanic eruptions. \pengfeicom{Models are also expected to separately or simultaneously support three types of spatiotemporal scales: macroscopic scale, which may span decades, mesoscopic scale, which is aligned with real-world temporal and spatial dynamics, and microscopic scale, which has millisecond-level precision.  }

     \end{itemize}

\subsection{Definition of Navigation Mode}
    \xintaocom{To better characterize how a video generation model functions as a higher generation world model, particularly in terms of its interaction with and response to external signals, we introduce the concept of navigation mode. This formulation provides a principled framework for understanding how external guidance signals condition the generative process within a world model. It also enables a clearer theoretical distinction between the spatial conditioning methods and navigation conditioning methods, which differ in their degrees of environmental interactivity and control flexibility.}
    
    We formally define a navigation mode as a structured interface through which an external condition signal guides the generative process. As shown in the Table~\ref{tab:condition_navigation_comparison}, a condition signal is considered a valid instance of navigation mode rather than a spatial condition only if it satisfies a triad of essential properties. We represent this triplet as $\{T, R, S\}$, capturing the fundamental requirements for temporality, content independence, and spatial reasoning.
    
    T - Temporality: The navigation mode must be defined as a temporally ordered sequence or influence the whole duration. This ensures that the guidance signal evolves over time, reflecting realistic changes in intent, observation, or control.
    
    R – Content Independence: The navigation mode must not explicitly reference the content and spatial characteristics within the video, including semantic maps, layouts, textual descriptions, motion poses, or depth maps. These types of conditions anchor the generation to specific, interpretable objectives and therefore require pairing with the original video content. They cannot be freely transferred to video generation in other background contexts, which inherently limits their interactive capability. Thus, we define the navigation mode as the content independence type of condition.

    S – Spatial Reasoning: The navigation mode must support spatial reasoning across the generated sequence. This implies that the world model must understand not just static spatial layouts but also dynamic transformations (\eg, agent motion, object displacement) to effectively fulfill the navigation mode.

\begin{table*}[t]
\resizebox{\textwidth}{!}{%
\centering
\small
\setlength{\tabcolsep}{4.5pt} 
\begin{tabular}{c|ccccccc|ccc}
\toprule
\multirow{2}{*}{Characteristics} & \multicolumn{7}{c|}{Spatial Condition} & \multicolumn{3}{c}{Navigation Mode} \\

& HD Map & Layout & Text Description & Canny & Depth & Sketch & Motion Pose & Trajectory & Action & Text Instruction\\
\midrule
Temporality & \xmark & \xmark & \xmark & \cmark & \cmark & \cmark & \cmark & \cmark & \cmark & \cmark \\

Content Independence & \xmark & \xmark & \cmark & \xmark & \xmark & \xmark & \xmark & \cmark & \cmark & \cmark \\

Spatial-Reasoning & \cmark & \cmark & \xmark & \xmark & \cmark & \cmark & \xmark & \cmark & \cmark & \cmark \\

\bottomrule
\end{tabular}
}
\vspace{0.1cm}
\caption{\textbf{Comparison Between Spatial Conditions and Navigation Modes.} In this table, we compare the key terms “spatial condition” and “navigation mode”, focusing on their main characteristics and differences. The comparison is conducted primarily across three dimensions: temporality, content independence, and spatial reasoning.}
\label{tab:condition_navigation_comparison}
\end{table*}

    
    Only when all three criteria are satisfied can a condition be said to activate the navigation mode of a world model. This triadic formulation provides a systematic way to assess whether a video generation model exhibits genuine planning and interactiveness, as opposed to merely replicating appearance or motion patterns. As such, the navigation mode serves as a litmus test for the maturity of world model based generation systems in both controlled and interactive settings. \xintaocom{In addition, future applications such as VR/AR or embodied AI may unify navigation and interaction via shared state feedback loops, extending beyond the current taxonomy.}

\section{Roadmap}
\label{sec: roadmap}

As discussed in Section~\ref{sec:introduction}, video generation models fundamentally comprise two components: an implicit world model and a video renderer. The world model is responsible for simulating and predicting the evolution of world states, encapsulating physical laws, causal dynamics, flexible controllability, and real-world planning. The video renderer, in turn, translates the internal states of the world model into visual outputs, specifically in video format, interpretable and accessible to both humans and intelligent agents.

This formulation underscores that video generation is not solely about producing realistic visuals; rather, it is about simulating and visualizing coherent world dynamics. As video generation increasingly assumes the role of a world model, its development can be characterized by four major generations.

\begin{figure}[!t]
      \centering
      \includegraphics[width=\linewidth]{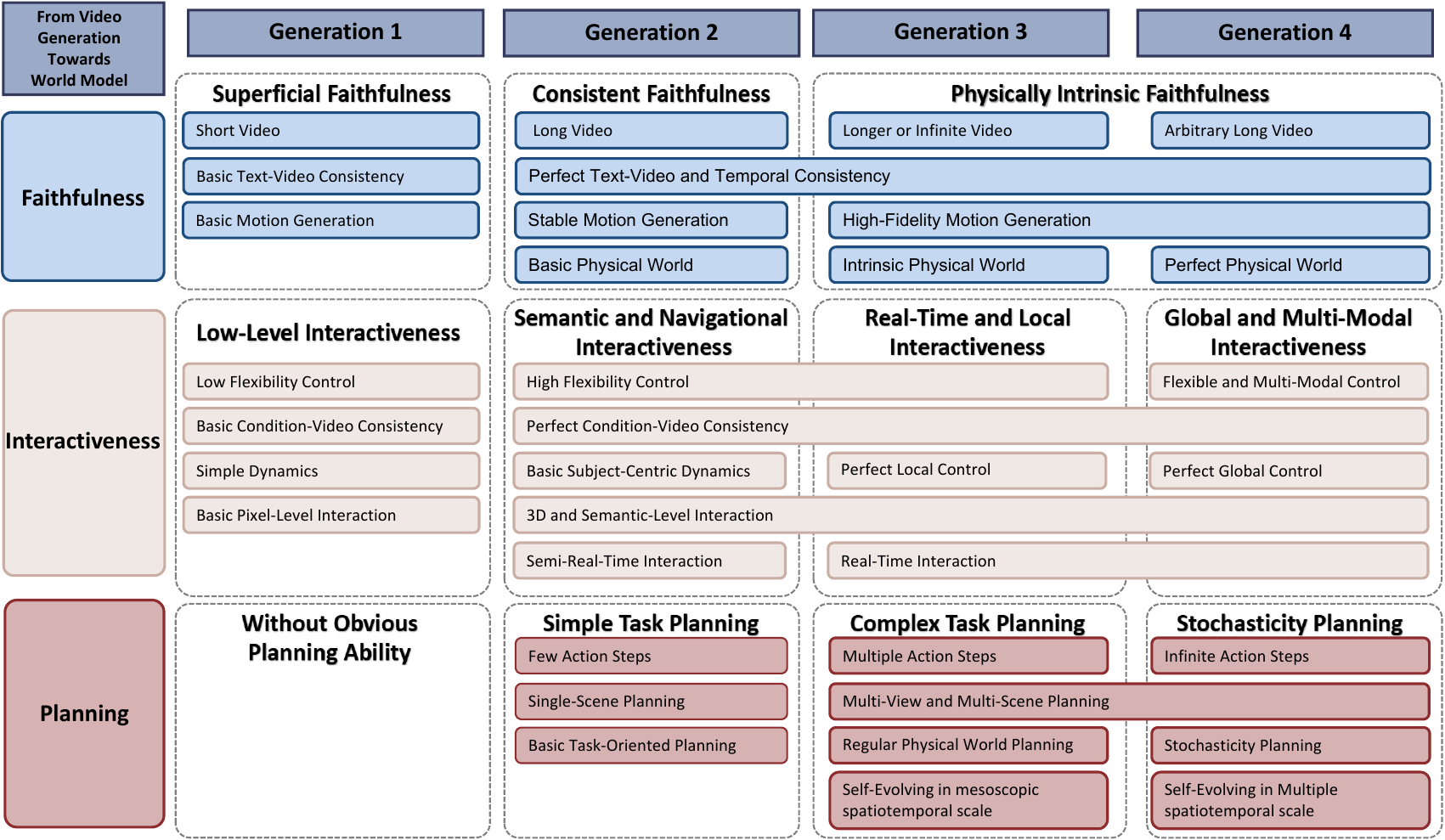}
      \caption{\textbf{Overview of the Capabilities of World Model Across 4 Generations.} This figure presents the three main capabilities of world models, along with their corresponding secondary capabilities under each category.}
      \label{fig:generation_detail}
    \end{figure}

Figure~\ref{fig:generation_detail} outlines the expected levels of faithfulness, interactivity, and planning across these generations, along with the corresponding advancements in capability at each stage. This figure further decomposes these three core dimensions into more fine-grained second- and third-level sub-capabilities. The characteristics and distinctions of the four generations are elaborated below:

\noindent    
\textbf{(1) Generation 1 - Faithfulness: Superficial Simulation of the Real World}

\noindent \textit{\textbf{Superficial Faithfulness.}} Faithfulness is the primary capability emphasized in Generation 1, as it represents the foundational step from video generation towards world models, which shifts from static realism to realistic motion dynamics.
This generation is categorized into two stages: the Basic Level and the Advanced Level.
The Basic Level consists of traditional and classical approaches, primarily based on GANs~\cite{liu2021generative} and early diffusion models.
In contrast, the Advanced Level encompasses video generation models that exhibit superficial faithfulness and low-level interactiveness, while still lacking explicit planning capabilities. 

\xintaocom{In this axis, short and long faithfulness denote the temporal span over which realistic and physically consistent generations can be maintained. We adopt a practical threshold where long corresponds to sequences of five seconds or longer at 24 FPS, following the convention that such durations begin to reveal a model’s stability as a world simulator rather than a mere simple video generator or single motion synthesizer. This temporal coherence implicitly depends on the model’s internal memory mechanisms, even though memory itself is not presented as a separate capability axis. Thus, although memory~\cite{po2025long} is important capability to the operation of any world model, we treat it as an underlying mechanism that enhances along with the mentioned observable capabilities, such as temporal faithfulness and consistent planning rather than as a standalone dimension in this taxonomy.}


As illustrated in Figure~\ref{fig:generation_detail}, models in Generation 1 are expected to achieve superficial faithfulness, typically demonstrated by the ability to generate short video clips, such as animated video, with tolerable motion distortions.
In addition, these models are required to maintain basic video-text consistency, accurately reflecting the primary subject and general background content. However, they often miss specific entities, distort style ordering, and produce unrealistic or incoherent motion patterns at a relatively high frequency.

\noindent \textit{\textbf{Low-Level Interactiveness.}} In terms of interactiveness, Generation 1 models exhibit low-level interactiveness, enabling pixel-level interaction and limited controllability. These systems typically support only a single short action composed of a few steps, guided by simple commands such as ``jump," ``grab the cup," or ``turn left." In contrast, they fail to interpret or execute more complex instructions such as ``clean the table," ``cook dinner," or ``find the key in several drawers."
Additionally, these models achieve basic condition-to-video consistency, where the generated motion aligns with input conditions to some extent but is often accompanied by subject distortions and incoherent background transitions.
While some approaches may incorporate auxiliary conditioning inputs, they primarily rely on spatial modalities, such as sketches, layouts, depth maps, and segmentation masks, which results in limited flexible controllability.\\
\noindent\textit{\textbf{Without Obvious Planning Ability.}} Due to the restricted faithfulness and interactiveness in Generation 1, planning capability has not yet emerged at this stage.

\noindent
\textbf{(2) Generation 2 - Interactiveness: Controllable and Interactive Simulation of the Real World}

\noindent \textit{\textbf{Semantic and Navigational Interactiveness.}} Interactiveness is the core capability emphasized in Generation 2, termed as semantic and navigational interactiveness, marking a significant advancement in the dimensions of control flexibility, condition-video consistency, subject-centric controllability, and intrinsic model competence. Models in this generation enable flexible control, particularly conditioned on navigation modes, including actions, text commands, and predefined trajectories. They also exhibit emerging basic reasoning abilities, such as inferring motion sequences from high-level commands. In terms of condition-video consistency, Generation 2 models demonstrate notable improvements: they can generate complete and coherent motions with minimal dynamic distortion, and maintain visually appropriate and semantically aligned backgrounds. Another hallmark of this generation is basic subject-centric external control, where the model is capable of interpreting and executing control signals directed at a specific subject, such as instructing one agent to perform a sequence of actions or dynamically adjusting the viewpoint around that subject. However, these models typically operate within relatively static or simplified backgrounds. In addition, the intrinsic model capabilities have advanced significantly, as evidenced by the integration of 3D spatial understanding and stronger semantic comprehension rather than basic pixel-level understanding. Together, these advancements position Generation 2 as a transitional phase towards truly interactive and semantically grounded world modeling.

Additionally, the long-term vision of interactiveness is building a general-purpose simulator that models the real world and allows human or agent interaction. For example, simulation-based games such as Euro Truck Simulator or SimCity can be viewed as early domain-specific prototypes of interactive world models.

\noindent \textit{\textbf{Consistent Faithfulness.}} Beyond improved interactiveness, Generation 2 world models show significant progress in maintaining temporal coherence and video–text consistency across long and complex sequences. The ability to generate longer videos contributes directly to temporal consistency~\cite{wu2024freeinit}, ensuring stable object dynamics and scene layouts over time. Moreover, these models achieve perfect video-text consistency, faithfully rendering all mentioned entities, motions, and events aligned with the given inputs. This generation also begins to capture aspects of the basic physical world, including projective geometry and spatial appropriateness, which contribute to more physically plausible and coherent video generations.

\noindent \textit{\textbf{Simple Task Planning.}} Additionally, this generation marks the emergence of simple task planning capabilities. 
While still limited in scope, these models \pengfeicom{show early signs of task-oriented planning, such as generating video content that follows a coherent intention or directive within approximate ten steps, albeit without physical evolution or complex interaction. They can handle basic short-horizon goal-directed behaviors observable in certain video and robot application models, such as those navigating via a goal image or executing two-step actions (\eg, pick-and-place). This reflects an early form of planning competence, laying the groundwork for advanced decision-making and long-horizon reasoning in subsequent generations and downstream tasks.}


\noindent
\textbf{(3) Generation 3 - Planning: Real-Time and Complex Prediction of the Real World}

\noindent \textit{\textbf{Complex Task Planning.}} Planning refers to the model’s ability to simulate the future evolution of a given world state. \pengfeicom{Generation 3 world models take this capability to a new level, achieving complex task planning. This includes generating long-term video sequences that exhibit self-evolving progress at a mesoscopic spatiotemporal scale, enabling the simulation of complex tasks with dozens or even hundreds of motion steps involving multiple interacting entities, dynamic viewpoint transitions, and scene transformations. These planning outcomes are not static; they can adapt in real time to interactions from both the internal state and the external environment.}

The broader vision for this level of planning is to faithfully simulate the evolution of the physical world under complex systems, such as weather patterns, narrative plots, cooking processes in robotics, population dynamics, or animal migrations. A vivid imagination of this capability is portrayed in Liu Cixin’s sci-fi novel The Mirror, where a “super simulator” is capable of projecting the future of the world with arbitrary precision, not merely replaying the past, but modeling the living, ever-changing future.

\noindent \textit{\textbf{Physically Intrinsic Faithfulness.}} Beyond planning, Generation 3 world models reach physically intrinsic faithfulness, that is, a new pinnacle in physical plausibility, enabling simulations that evolve according to the intrinsic physical principles of the real world. The models are capable of generating arbitrarily long video sequences, which brings higher complexity and enables them to create new motions, entities, viewpoints, and scenes over time, all while maintaining temporal coherence.
More impressively, these models internalize the laws of physics themselves. Generation 3 models show evidence of learning causal dynamics across multiple physical domains. Examples include rigid-body mechanics (\eg, free-fall, collisions), fluid dynamics, and potentially electromagnetic effects. This represents a fundamental leap: instead of approximating appearances, these models simulate underlying causal processes, leading to greater scientific fidelity and application potential.

\noindent \textit{\textbf{Real-Time and Local Interactiveness.}} In terms of interaction, Generation 3 achieves real-time and local interactiveness with high-fidelity controllability, enabling frame-level interaction without perceptible delay. Users can engage with the world model seamlessly, issuing commands and stimuli that lead to instant, coherent changes, whether it's modifying an object's trajectory, switching perspectives, or inserting a new entity into the scene.
In addition, local control becomes precise and expressive. These models support subject-centric manipulation with fine-grained attention to contextual and background consistency. For example, a user can focus on a single character’s behavior while the surrounding environment continues to evolve naturally with rich, photorealistic details, all without compromising visual or physical consistency.

\noindent
\textbf{(4) Generation 4 - Stochasticity: Low-Probability and Multi-Scale Modeling of the Real World}

\noindent \textit{\textbf{Stochasticity Planning.}} Generation 4 world models advance planning capabilities by incorporating stochasticity-aware reasoning, enabling the simulation of both high-probability and low-probability events aligned to the real-world distribution. This supports not only deterministic future prediction but also probabilistic modeling of diverse potential outcomes, especially proactive modeling of black swan events such as earthquakes, tsunamis, financial crises, and asteroid impacts.

Furthermore, Generation 4 achieves arbitrary spatial and temporal scale planning. In the spatial domain, the model can plan across macroscopic scales such as universe-level evolution and microscopic scales like microbial dynamics or atomic-level transitions. Similarly, in the temporal domain, the model is capable of operating across vast time scales, from long-term evolution spanning years or centuries (requiring time compression and critical event selection ability), to mid-scale physical world dynamics, down to fine-grained, high-frequency phenomena such as insect wing beats or human pupil micro-movements. This ability to plan across stochastic events and arbitrary scales represents a critical step towards building general-purpose simulation engines that align more closely with the complexity and uncertainty of the real world.

 Some recent works~\cite{nwm, aether, assran2025vjepa2} have begun to explore the mesoscopic-scale planning capabilities of world models, but both the microscopic and macroscopic scales remain largely underexplored. However, these scales are essential for faithfully simulating the physical world, as they capture fine-grained interactions and high-level structural dynamics, respectively.

 \noindent \textit{\textbf{Physically Intrinsic Faithfulness.}} In Generation 4, the capacity for physical faithfulness remains consistent with that of Generation 3. This is because Generation 3 has already achieved a high level of physical realism by accurately adhering to intrinsic physical laws and simulating plausible, causally coherent environments. As such, Generation 4 inherits and maintains this state-of-the-art fidelity, providing a solid and reliable physical foundation upon which more advanced capabilities, such as stochasticity-aware planning and global interactiveness, are built.

 \noindent \textit{\textbf{Global and Multi-Modal Interactiveness.}} In addition to its advanced planning capabilities, Generation 4 world models also exhibit a leap in global and multi-modal interactiveness. These models are capable of predicting long-term, multimodal influences resulting from external interventions, allowing for sustained, temporally extended interactions across vision, language, and control modalities. At the core of this interactiveness is a form of global control, where an internal agent equipped with a mental world model acts as the primary decision-making entity within the simulated environment. Moreover, these models support multi-entity control by responding to external signals, enabling coordination among multiple agents or systems within the scene. The incorporation of dynamic and evolving backgrounds further enriches the simulation, allowing for more realistic and adaptive world modeling.

 \pengfeicom{To concretely distinguish the capabilities of Generations 1–4, consider a consistent scenario: a person making a cup of coffee in a kitchen.}
    
    \begin{itemize}
    
    \item \pengfeicom{Generation 1 models can generate a few frames of water pouring or a static view of a coffee cup appearing, with no awareness of human intent or task continuity.}
    
    \item \pengfeicom{Generation 2 models can depict a simple, goal-directed action sequence within a short horizon, such as a person picking up a kettle and pouring water into a cup following short term, simple commands such as “grab the cup" and "pour the water into the cup." The motion is locally consistent, and the action follows a visible goal, but the model lacks a persistent understanding of the broader task or multi-step dependencies.}
    
    \item \pengfeicom{Generation 3 models begin to exhibit real-time generation and interactive consistency, enabling video generation adaptively to a abstract and long-horizon command such as “make a cup of coffee.” They maintain spatial coherence of the kitchen layout and object positions over several seconds, showing flexible navigation and environmental awareness.}
    
    \item \pengfeicom{Generation 4 models are envisioned to autonomously complete the entire coffee-making procedure, planning and executing multi-step actions such as heating water, grinding beans, pouring, and serving, while maintaining physical faithfulness, temporal continuity, and coherent interaction with dynamic objects in the scene. Moreover, through multiple inference cycles, realistic incidents, such as accidentally spilling hot water onto the table, may emerge naturally, reflecting plausible real-world possibilities.}
    
    \end{itemize}
    \pengfeicom{This progressive depiction within the same environment clarifies how each generation expands its prediction and planning ability.}

\section{Methods: A Hierarchical Taxonomy}
\label{sec:methods}

\xintaocom{Before delving into each generation in detail, it is important to emphasize that the three core capabilities, faithfulness, interactiveness, and planning, develop in parallel rather than appearing sequentially. Each generation highlights one capability as its dominant focus, reflecting the primary research emphasis and maturity level of that stage, yet all three capabilities continue to advance simultaneously across generations. This co-evolution reflects the natural progression of video generation models towards comprehensive world modeling, where visual realism, controllability, and long-horizon reasoning are continuously refined in tandem rather than introduced one after another.}

\xintaocom{Similarly, the application domains, general scenes, robotics, autonomous driving, and gaming, analyzed in this section are not exclusive to any single generation. Instead, they overlap across all four, as the underlying modeling principles remain consistent while the level of capability and control gradually increases. We organize these applications under Generation 2 primarily because most of the recent works and open-source systems lie in this stage. Subsequent generations naturally extend these same applications towards higher-level planning, real-time interaction, and stochastic world simulation.}

\begin{figure}[!t]
      \centering
      \includegraphics[width=\linewidth]{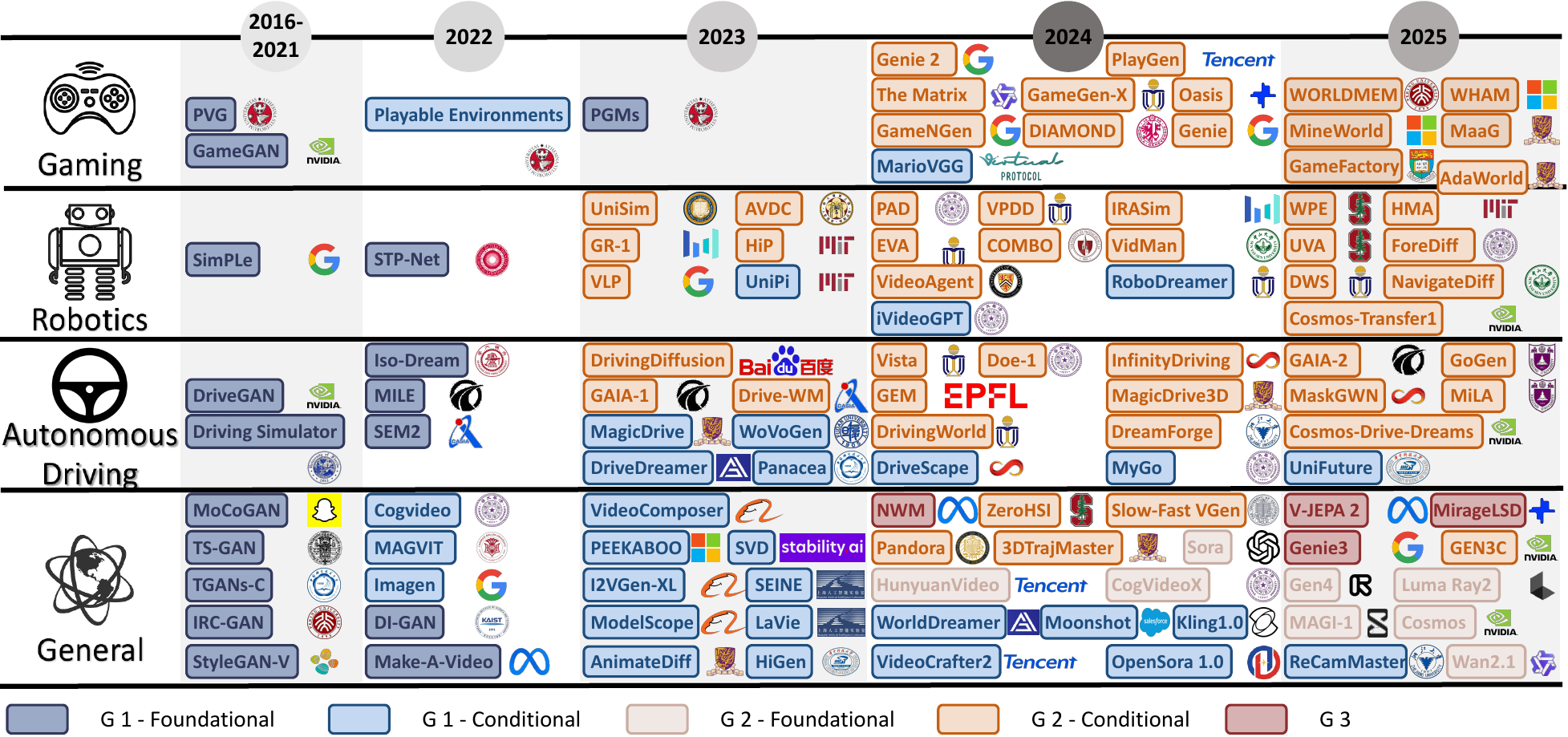}
      \caption{\textbf{Chronological Overview of Methods from Video Generation to World Models.} The figure presents a chronological overview along the horizontal axis, categorizing existing methods by four application domains along the vertical axis: general scenes, robotics, autonomous driving, and gaming. Additionally, different colors are used to indicate different generations of world models. }
      \label{fig:main_methods}
    \end{figure}


\definecolor{crText}{RGB}{243, 240, 161}    
\definecolor{crImage}{RGB}{247, 173, 190}   
\definecolor{crVideo}{RGB}{191, 155, 203}   

\definecolor{cr3D}{RGB}{139, 218, 225}   
\definecolor{crGeometry}{RGB}{255, 200, 196} 
\definecolor{crPhysics}{RGB}{197, 225, 222} 
\definecolor{crMotionV}{RGB}{219, 210, 201}   

\definecolor{crNone}{RGB}{238, 237, 234}    
\definecolor{crTraj}{RGB}{250, 234, 219}
\definecolor{crCameraM}{RGB}{162, 180, 156}  

\definecolor{crAction}{RGB}{240, 212, 209}  
\definecolor{crGoal}{RGB}{202, 210, 219}  
\definecolor{crVolume}{RGB}{160, 215, 209}  
\definecolor{crDepth}{RGB}{112, 206, 234}   
\definecolor{crLayout}{RGB}{240, 179, 210} 
\definecolor{crBBox}{RGB}{239, 208, 151}    

\begin{table*}[!t]
    \setlength\extrarowheight{1pt}
    \centering
    \caption{\pengfeicom{\textbf{Summary and Comparison of Works for Video Foundation Models.} * indicates models that include audio input or output. We compare models in terms of their backbone architectures, supported frame counts, and output resolutions. For models released in multiple duration or resolution versions, we report the highest available configuration by default.}}
    \label{tab:Foundation}
    {\fontsize{7pt}{11pt}\selectfont
    \begin{tabularx}{\linewidth}{p{2.6cm}p{1.2cm}p{1cm}p{1.5cm}||p{2.8cm}p{1.2cm}p{1cm}p{1.5cm}}
    \cmidrule{1-8}
    
       \bf{Method} 
       & \bf{Backbone}   
       & \bf{Frames}
       & \bf{Resolution}
       & \bf{Method} 
       & \bf{Backbone}   
       & \bf{Frames}
       & \bf{Resolution}
    \\
    \cmidrule{1-8} 
    \gray%
        \multicolumn{8}{c}{Generation 1 World Model} \\
    \cmidrule{1-8}
        LTX-Video \cite{hacohen2024ltxvideo}
        & DiT
        & 121        
        & $768 \times 512$
        & STIV \cite{lin2024stiv}
        & DiT
        & 60
        & $512 \times 512$
        \\
        JT-CV \cite{jvcvt2v}
        & -
        & 51         
        & $1214 \times 2158$
        & Hailuo 01 \cite{hailuo24minmax}
        & -
        & 144
        & $1280 \times 720$
        \\
        OpenSora 1.3 \cite{zheng2024opensora}
        & DiT
        & 113        
        & $1280 \times 720$
        & Open-Sora Plan \cite{lin2024opensoraplan}
        & DiT
        & 121
        & $1024 \times 576$
        \\
        EasyAnimate V5 \cite{xu2024easyanimate}
        & DiT
        & 49         
        & $1024 \times 1024$
        & King 1.0 \cite{klingai2024kling}
        & DiT
        & 300
        & $1920 \times 1080$
        \\
        Lumiere~\cite{bar2024lumiere}
        & UNet
        & 80         
        & $1024 \times 1024$
        & VideoCrafter2 \cite{chen2024videocrafter2}
        & UNet
        & 16
        & $512 \times 320$
        \\
        Latte \cite{ma2024latte}
        & DiT
        & 16      
        & $512 \times 512$
        & MagicVideo-V2 \cite{wang2024magicvideov2}
        & UNet
        & 94
        & $1048 \times 1048$
        \\
        Mira \cite{ju2024miradata}
        & DiT
        & 2880         
        & $384 \times 240$
        & HiGen \cite{qing2024higen}
        & UNet
        & 32
        & $448 \times 256$
        \\
        SVD \cite{blattmann2023svd}
        & UNet
        & 25     
        & $1024 \times 576$
        & I2VGen-XL \cite{zhang2023i2vgen}
        & UNet
        & -
        & $1280 \times 720$
        \\
        SEINE \cite{chen2023seine}
        & UNet
        & 16       
        & $560 \times 240$
        &VideoCrafter1 \cite{chen2023videocrafter1}
        & UNet
        & 16
        & $1024 \times 576$
        \\
        DynamiCrafter \cite{xing2024dynamicrafter}
        & UNet
        & 16         
        & $1024 \times 576$
        & MAGVIT-v2 \cite{yu2023magvitv2}
        & ViT
        & -
        & $640 \times 360$
        \\
        Show-1 \cite{zhang2024show1}
        & UNet
        & 29     
        & $576 \times 320$
        & LaVie~\cite{wang2025lavie}
        & UNet
        & 16
        & $512 \times 512$
        \\
        ModelScope~\cite{wang2023modelscope}
        & UNet
        & 16
        & $256 \times 256$
        &AnimateDiff v2~\cite{guo2023animatediff}
        & UNet
        & 16     
        & $512 \times 512$
        \\
        Pika 1.5~\cite{pika23}
        & - 
        & 120
        & $1280 \times 720$
        &Gen-2~\cite{gen223}
        & -
        & 96         
        & $1408 \times 768$
        \\
        MAGVIT~\cite{yu2023magvit}
        & UNet
        & 192         
        & $128 \times 128$
        & MagicVideo~\cite{zhou2022magicvideo}
        & UNet
        & -
        & $1024 \times 1024$
        \\
        Imagen Video~\cite{Ho22imagenvideo}
        & UNet
        & 128         
        & $1280 \times 768$
        & Make-A-Video~\cite{singer2022makeavideo}
        & UNet
        & -
        & $768 \times 768$
        \\
        CogVideo~\cite{hong2022cogvideo}
        & DiT
        & 32      
        & $480 \times 480$
        & RepVideo~\cite{si2025repvideo}
        & -
        & 16
        & $512 \times 512$
        \\

        InstructVideo~\cite{yuan2024instructvideo}
        & UNet
        & 16    
        & $256 \times 256$
        & Emu Video~\cite{girdhar2023emuvideo}
        & UNet
        & 16
        & $512 \times 512$
        \\

        Allegro~\cite{zhou2024allegro}
        & DiT
        & 88
        & $1280 \times 720$
        & DimensionX~\cite{sun2024dimensionx}
        & AR
        & 16    
        & $256 \times 256$
        \\
        StreamingT2V~\cite{henschel2025streamingt2v}
        & AR
        & 1200
        & $256 \times 256$
        & VideoPoet~\cite{kondratyuk2023videopoet}
        & AR
        & 80    
        & $512 \times 896$
        \\
        VideoTetris~\cite{tian2024videotetris}
        & UNet
        & 16 
        & $512 \times 320$
        & Owl-1~\cite{huang2024owl}
        & UNet
        & -
        & -
        \\
        
    \cmidrule{1-8} 
    \gray%
        \multicolumn{8}{c}{Generation 2 World Model} \\
    \cmidrule{1-8}
    
        Hailuo 02~\cite{hailuo0225minmax}
        & -
        & 240
        & $1920 \times1080$
        & Seedance 1.0~\cite{gao2025seedance}
        & DiT
        & 240
        & $1920 \times1080$
        \\
        MAGI-1~\cite{teng2025magi}
        & AR-DiT
        & 96
        & $720 \times 720$
        & \pengfeicom{\jiamingcom{Veo 3*}~\cite{Veo325}}
        & -
        & 192
        & $1280 \times 720$
        \\
        Nova Reel~\cite{nova25amazonaws}
        & -
        & 2880
        & $1280 \times 720$
        & Gen-4~\cite{runway2025gen4}
        & -
        & 240
        & $1280 \times 720$
        \\
        Wan 2.1~\cite{wan2025wan}
        & DiT
        & 81
        & $1280 \times 720$
        & Step-Video-T2V~\cite{ma2025stepvideo}
        & -
        & 200
        & $960 \times 540$
        \\
        Open-Sora 2.0~\cite{peng2025opensora2}
        & DiT
        & 120
        & $1024 \times 576$
        & SkyReels-V2~\cite{chen2025skyreels}
        & AR-DiT
        & 121
        & $1280 \times 720$
        \\
        MiracleVision~\cite{MiracleVision25meitu}
        & -
        & 120
        & $720 \times 480$
        & PixverseV4.5~\cite{pixverse2025}
        & -
        & 192
        & $1920 \times 1080$
        \\
        Vchitect-2.0~\cite{fan2025vchitect}
        & -
        & 40
        & $768 \times 432$
        & Cosmos-Predict2~\cite{agarwal2025cosmos}
        & DiT/AR
        & 80
        & $1280 \times 720$
        \\
         HunyuanVideo~\cite{kong2024hunyuanvideo}
        & DiT
        & 129
        & $1280 \times 720$
        & APT2~\cite{lin2025apt2}
        & AR-DiT
        & 1440        
        & $1280 \times 720$
        \\
        Mochi-1~\cite{genmo2024mochi1}
        & DiT
        & 150
        & $1969 \times 960$
        &CogVideoX1.5~\cite{yang2024cogvideox}
        & DiT
        & 160
        & $1360 \times 768$
        \\
        \pengfeicom{\jiamingcom{Gen-3*~\cite{runway2024gen3}}}
        & -
        & 240
        & $1280 \times 720$
        & Kling 2.1 Master~\cite{klingai2024kling}
        & DiT
        & 240
        & $1920 \times 1080$
        \\
        EasyAnimateV5.1~\cite{xu2024easyanimate}
        & DiT
        & 49
        & $1024 \times 1024$
        & Jimeng~\cite{jimengai24jimeng}
        & -
        & 96
        & $1280 \times 720$
        \\
        Vidu Q1~\cite{bao2024vidu}
        & UNet
        & 120
        & $1920 \times 1080$
        &Sora~\cite{sora}
        & DiT
        & 480
        & $1920 \times 1080$
        \\
        Luma Ray2~\cite{luma24}
        & -
        & 240        
        & $1920 \times 1080$
        & 
        \pengfeicom{\jiamingcom{Pika 2.2*}~\cite{pika23}}
        & -
        & 240
        & $1920 \times 1080$
        \\

        CausVid~\cite{yin2025causvid}
        & AR-DiT
        & 121       
        & $640 \times 352$
        & Self Forcing~\cite{huang2025selfforcing}
        & AR-DiT
        & 81
        & $854 \times 480$
        \\
        Movie Gen~\cite{polyak2024moviegen}
        & DiT
        & 250   
        & $1920 \times 1080$
        &\pengfeicom{\jiamingcom{Imagine v0.9*~\cite{imaginegrok}}}
        & - 
        & 80
        & $464 \times 688$
        \\
        \pengfeicom{\jiamingcom{Wan 2.2*~\cite{wan22}}}
        & DiT
        & $120$
        & $1280 \times 720$
        &\pengfeicom{\jiamingcom{Sora 2*~\cite{openai2025sora2}}}
        & DiT
        & $1800$
        & $1920 \times 1080$\\
        \pengfeicom{\jiamingcom{Luma Ray3*~\cite{luma2025ray3}}}
        & -
        & $240$
        & $3840 \times 2160$
        &\pengfeicom{\jiamingcom{Vibes*~\cite{meta2025vibes}}}
        & -
        & $240$
        & -
        \\
        Kling-Omni~\cite{team2025kling}
        & DiT
        & $240$
        & $1920 \times 1080$
        &SemanticGen~\cite{bai2025semanticgen}
        & DiT
        & $1440$
        & $832 \times 480$
        \\
        MemFlow~\cite{ji2025memflow}
        & AR-DiT
        & $1122$
        & $832 \times 480$
        \\

    \cmidrule{1-8} 
    \gray%
        \multicolumn{8}{c}{Generation 3 World Model} \\
    \cmidrule{1-8}

    V-JEPA 2~\cite{assran2025vjepa2}
    & AR
    & -
    & -
    & MirageLSD~\cite{mirage2025}
    & AR-DiT
    & Infinite
    & $512 \times 512$
\\
    Genie 3~\cite{google2025genie3}
    & AR
    & $>1440$
    & $1280 \times 720$
    & Magica 2~\cite{magica22025}
    & AR-DiT
    & Minutes
    & -
\\
    \cmidrule{1-8}
    \end{tabularx}
    }

\end{table*}



\definecolor{crText}{RGB}{243, 240, 161}    
\definecolor{crImage}{RGB}{247, 173, 190}   
\definecolor{crVideo}{RGB}{191, 155, 203}   

\definecolor{cr3D}{RGB}{139, 218, 225}   
\definecolor{crGeometry}{RGB}{255, 200, 196} 
\definecolor{crPhysics}{RGB}{197, 225, 222} 
\definecolor{crMotionV}{RGB}{219, 210, 201}   

\definecolor{crNone}{RGB}{238, 237, 234}    
\definecolor{crTraj}{RGB}{250, 234, 219}
\definecolor{crCameraM}{RGB}{162, 180, 156}  

\definecolor{crAction}{RGB}{240, 212, 209}  
\definecolor{crGoal}{RGB}{202, 210, 219}  
\definecolor{crVolume}{RGB}{160, 215, 209}  
\definecolor{crDepth}{RGB}{112, 206, 234}   
\definecolor{crLayout}{RGB}{240, 179, 210} 
\definecolor{crBBox}{RGB}{239, 208, 151}    

\begin{table*}[!t]
    \setlength\extrarowheight{1pt}
    \centering
    \caption{\textbf{Summary and Comparison of Works for Video Generation as World Model in General Scene}. 
    }
    \label{tab:General-methods}
    {\fontsize{5pt}{7pt}\selectfont
    \begin{tabularx}{\linewidth}{p{2.8cm}p{1.5cm}p{1.8cm}p{1.9cm}p{3.5cm}X}
    \toprule
       \bf{Method} 
       & \bf{Condition}
       & \bf{Navigation}
       & \bf{Output}   
       & \bf{Task} 
       & \bf{Foundation Model} \\
    \midrule 
    \gray%
    \multicolumn{6}{c}{Generation 1 World Model} \\
    \midrule
        
        Moonshot~\cite{zhang2024moonshot} 
        & Geometry
        & None          
        & Video
        & Generation, Editing
        & -\\
        
        SparseCtrl~\cite{guo2024sparsectrl} 
        & Geometry
        & None    
        & Video
        & Gen. Ani.
        &  AnimateDiff~\cite{guo2023animatediff}  \\

        ConditionVideo~\cite{peng2024conditionvideo} 
        & Geometry
        & None           
        & Video
        & Generation 
        & -  \\
        
        ControlVideo~\cite{zhang2023controlvideo}
        & Geometry
        & None       
        & Video
        & Generation
        & -  \\

        VideoComposer~\cite{wang2023videocomposer}
        & Geometry
        & Trajectory 
        & Video
        & Generation, Inpainting
        & - \\

        Text2Video-Zero~\cite{khachatryan2023text2video}
        & Geometry
        & None 
        & Video
        & Generation, Inpainting
        & -  \\

        Diffusion4D ~\cite{liang2024diffusion4d}
        & 3D
        & None       
        & Video, 3D
        & Generation, Construction
        & ModelScope~\cite{wang2023modelscope} \\

        & 3D
        & None   
        & Video, 3D
        & Generation, Construction
        & I2VGen-XL~\cite{zhang2023i2vgen}  \\
        
        SV3D~\cite{voleti2024sv3d}
        & 3D
        & None   
        & Video, 3D
        & Generation
        & SVD~\cite{blattmann2023svd}  \\
        
        V3D~\cite{chen2024v3d} 
        & 3D
        & None   
        & Video, 3D
        & Generation, Construction
        & SVD~\cite{blattmann2023svd}  \\

        VideoMV~\cite{zuo2024videomv}
        & 3D
        & None
        & Video, 3D
        & Generation, Construction
        & ModelScope\cite{wang2023modelscope} \\

        & 3D
        & None
        & Video, 3D
        & Generation, Construction
        & I2VGen-XL\cite{zhang2023i2vgen}  \\

        PhysGen~\cite{liu2024physgen} 
        & Physics
        & None
        & Video
        & Generation
        & SEINE~\cite{chen2023seine}  \\

        Go-with-the-Flow~\cite{burgert2025gowiththeflow}
        & 3D
        & Trajectory       
        & Video
        & Motion Control, Camera Control
        & CogVideoX~\cite{yang2024cogvideox}  \\

        Motion Prompting~\cite{geng2025motionprompting}
        & None
        & Trajectory       
        & Video
        & Motion Control, Camera Control
        & Lumiere~\cite{bar2024lumiere}  \\

        SG-I2V~\cite{namekata2024sgi2v}
        & None
        & Trajectory    
        & Video
        & Motion Control, Camera Control
        & SVD~\cite{blattmann2023svd}  \\

        Image Conductor~\cite{li2025imageconductor}
        & None
        & Trajectory      
        & Video
        & Motion Control, Camera Control
        & Animatediff~\cite{guo2023animatediff} \\
        
        TrailBlazer~\cite{ma2024trailblazer} 
        & None
        & Trajectory      
        & Video
        & Motion Control 
        & ModelScope~\cite{wang2023modelscope}  \\

        MotionCtrl~\cite{wang2024motionctrl} 
        & None
        & Trajectory       
        & Video
        & Motion Control, Camera Control
        & VideoCrafter1~\cite{chen2023videocrafter1}  \\

        DragAnything~\cite{wu2024draganything} 
        & Geometry
        & Trajectory       
        & Video
        & Motion Control, Camera Control
        & SVD~\cite{blattmann2023svd}  \\
        
        PEEKABOO~\cite{jain2024peekaboo}
        & None
        & Trajectory     
        & Video
        & Motion Control 
        & ModelScope~\cite{wang2023modelscope}  \\

        FreeTraj~\cite{qiu2024freetraj} 
        & None
        & Trajectory        
        & Video
        & Motion Control 
        & VideoCrafter1~\cite{chen2023videocrafter1}  \\

        Direct-A-Video~\cite{yang2024directavideo}
        & None
        & Trajectory        
        & Video
        & Motion Control, Camera Control. 
        & ModelScope~\cite{wang2023modelscope}  \\

        FullDiT~\cite{ju2025fulldit} 
        & Geometry
        & Camera Motion    
        & Video
        & Generation
        & Private  \\
        
        ReCamMaster~\cite{bai2025recammaster} 
        & None
        & Camera Motion      
        & Video
        & Camera Control
        & -  \\

        CineMaster~\cite{wang2025cinemaster} 
        & 3D
        & Camera Motion
        & Video
        & Cameral Control
        & -  \\

        AC3D~\cite{bahmani2025ac3d}
        & None
        & Camera Motion 
        & Video
        & Camera Control
        & - \\

        VD3D~\cite{bahmani2024vd3d}
        & None
        & Camera Motion
        & Video
        & Camera Control
        & -  \\

        CameraCtrl~\cite{he2024cameractrl} 
        & None
        & Camera Motion 
        & Video
        & Camera Control
        & SVD~\cite{blattmann2023svd} \\

        & None
        & Camera Motion       
        & Video
        & Camera Control
        & AnimateDiff~\cite{guo2023animatediff}  \\

        GCD~\cite{van2024gcd} 
        & None
        & Camera Motion      
        & Video
        & Generation
        & SVD~\cite{blattmann2023svd}  \\

        CVD~\cite{kuang2024cvd}
        & None
        & Camera Motion   
        & Video
        & Camera Control
        & AnimateDiff~\cite{guo2023animatediff}  \\

        CamCo~\cite{xu2024camco}
        & None
        & Camera Motion     
        & Video
        & Camera Control
        & SVD~\cite{blattmann2023svd}  \\

    \midrule
    \gray%
    \multicolumn{6}{c}{Generation 2 World Model} \\
    \midrule

    SketchVideo~\cite{sketchvideo}
        & Geometry 
        & None           
        & Video
        & Generation, Editing
        & CogVideoX~\cite{yang2024cogvideox} \\

    DaS~\cite{das}
        & 3D
        & None    
        & Video
        & Motion Control, Camera Control, Motion Transfer, Animating
        & CogVideoX~\cite{yang2024cogvideox}  \\
        
    GS-DiT~\cite{bian2025gsdit} 
        & 3D
        & None     
        & Video
        & Generation
        & CogVideoX~\cite{yang2024cogvideox}  \\
    PISA~\cite{li2025pisa}
        & Physics
        & None   
        & Video
        & Generation
        & Open-Sora~\cite{zheng2024opensora} \\

    PhyT2V~\cite{xue2025phyt2v} 
        & Physics
        & None   
        & Video
        & Gen.
        & CogVideoX~\cite{yang2024cogvideox}  \\

    Zhao et al.~\cite{zhao2025synthetic}
        & Physics
        & None    
        & Video
        & Generation
        & MMDiT~\cite{esser2024mmdit}  \\

    WISA~\cite{wang2025wisa}  
        & Physics
        & None    
        & Video
        & Generation
        & CogVideoX~\cite{yang2024cogvideox}  \\

    CamCloneMaster~\cite{luo2025camclonemaster}
     & None
     & Camera Motion 
     & Video
     & Generation
     & Private
    \\
    Context as Memory~\cite{yu2025contextasmemory}
     & None
     & Camera Motion
     & Video
     & Generation
     & Private
    \\
    3DTrajMaster~\cite{fu20243dtrajmaster}
        & None
        & Trajectory, Camera Motion     
        & Video
        & Motion Control, Camera Control
        & Private \\

    CameraCtrl II~\cite{he2025cameractrl2} 
        & None
        & Camera Motion      
        & Video
        & Camera Control
        & - \\
    
    GEN3C~\cite{ren2025gen3c}
        & 3D
        & Camera Motion
        & Video
        & Camera Control
        &  SVD~\cite{blattmann2023svd}
        \\
    CamTrol~\cite{hou2024camtrol} 
        & 3D
        & Camera Motion       
        & Video
        & Camera Control
        & SVD~\cite{blattmann2023svd}  \\
        
    PhysDreamer~\cite{zhang2024physdreamer}
        & 3D
        & Action    
        & Video
        & Generation
        & -\\

    WonderPlay~\cite{li2025wonderplay}
        & None
        & Action    
        & Video
        & Generation
        & Go-with-the-Flow~\cite{burgert2025gowiththeflow}\\
    AETHER~\cite{aether}
        & None
        & Action    
        & Video, 3D, Action
        & Generation
        & CogVideo~\cite{hong2022cogvideo}\\

    SlowFast-VGen~\cite{hong2024slowfast}
        & None
        & Instruction      
        & Video
        & Generation
        & ModelScope~\cite{wang2023modelscope}  \\
        
     Pandora~\cite{xiang2024pandora}
        & None
        & Instruction      
        & Video
        & Generation
        & DynamiCrafter~\cite{xing2024dynamicrafter}  \\
    \midrule
    \gray%
    \multicolumn{6}{c}{Generation 3 World Model} \\
    \midrule

    NWM~\cite{nwm}
        & None
        & Action     
        & Video
        & Generation, Planning
        & - \\

    V-JEPA 2-AC~\cite{assran2025vjepa2}
        & None
        & Action    
        & Video
        & Generation, Planning
        & V-JEPA 2~\cite{assran2025vjepa2} \\

    MotionStream~\cite{shin2025motionstream}
        & None
        & Trajectory    
        & Video
        & Generation
        & Wan Series~\cite{wan2025wan, wan22}\\
    
    \bottomrule
    \end{tabularx}
    }
\end{table*}


\subsection{Generation 1 - Faithfulness: Accurate Simulation of the Real World}

\label{sec:method_G1}
As shown in Figure~\ref{fig:generation_detail}, in the first generation, the requirements for world models focus on the general video generation model capabilities and basic interaction characteristics. Similar to VBench series~\cite{huang2024vbench, huang2024vbench++}, the focus is placed on the superficial faithfulness, including pixel-level frame quality, the duration of the generated video, and video-text consistency. Building upon this foundation, we systematically categorize the existing methods in Generation 1 for the video foundation models~\ref{sec:method_general_foun}, spatial world models~\ref{sec:method_gen1_condition}, and navigated world models~\ref{sec:method_gen1_navigation}. 

\subsubsection{Foundation Models with Visual-Centric World Knowledge}
\label{sec:method_general_foun}
While video generation models in other domains~\cite{yuan2023physdiff} have also made significant progress, this survey focuses specifically on foundation models, pre-trained video generation methods that produce videos based solely on $I = \{T, O\}$ without relying on additional conditioning signals or modalities, due to their potential to support or even serve as comprehensive world models.
More importantly, the capabilities of video foundation models largely determine the effectiveness of subsequent adaptations or fine‑tuning for spatial conditions and navigation modes. 
With their foundational faithfulness and basic interactiveness, these models form the backbone upon which conditional mechanisms (\eg, ControlNet~\cite{zhang2023controlnet}, Multi-Modal Transformer, Cross-Attention, Concatenation, and Addition) can be applied to achieve high‑quality, physically plausible, and task‑aware video generation. 

In Generation 1, foundation models demonstrate superficial faithfulness and low-level interactiveness by generating short videos with basic text–video consistency and limited control over simple object motions. They capture key visual details such as object boundaries, textures, and coherent foreground and background layouts. These capabilities support world models in downstream applications like autonomous navigation~\cite{geng2025motionprompting, namekata2024sgi2v, ma2024trailblazer, wang2024motionctrl, wu2024draganything, jain2024peekaboo, qiu2024freetraj, yang2024directavideo, he2024cameractrl, kuang2024cvd} and interactive decision-making~\cite{wu2024ivideogpt, wang2024drivedreamer, wang2024drivewm, hu2022mile, zhu2025eotwm, micheli2022iris}.

We summarize representative pre-trained text-to-video (T2V) models~\cite{hong2022cogvideo, chen2023videocrafter1, chen2024videocrafter2, blattmann2023svd, wang2023modelscope, gen223, guo2023animatediff, zhang2024show1, wang2025lavie, qing2024higen, ma2024latte, wang2024magicvideov2, bar2024lumiere, klingai2024kling, henschel2025streamingt2v, girdhar2023emuvideo, kondratyuk2023videopoet, yuan2024instructvideo, zhou2024allegro, tian2024videotetris}, along with commonly used pre-trained image-to-video (I2V) models~\cite{zhang2023i2vgen, lin2024opensoraplan, hailuo24minmax, xu2024easyanimate, klingai2024kling, blattmann2023svd, yu2023magvitv2, chen2023videocrafter1, sun2024dimensionx, kondratyuk2023videopoet, guo2024i2vadapter, team2025kling}, examining their backbone architectures, video lengths, resolutions, and post-training capabilities related to interactivity and embodiment which are core abilities for serving as world models. This survey also covers both open-sourced models~\cite{ma2024latte, wang2025lavie, xu2024easyanimate, si2025repvideo, chen2024videocrafter2, guo2023animatediff, hacohen2024ltxvideo, chen2023videocrafter1, zheng2024opensora, zhang2024show1, lin2024opensoraplan, qing2024higen, wang2023modelscope, henschel2025streamingt2v, sun2024dimensionx, kondratyuk2023videopoet, zhou2024allegro} and commercial APIs~\cite{hailuo24minmax, klingai2024kling, gen223, pika23, girdhar2023emuvideo}. For instance, CogVideo~\cite{hong2022cogvideo} is widely regarded as an early open-source, large-scale pretrained T2V model that played a pivotal role in initiating the development of video foundation models. The release of Kling 1.0~\cite{klingai2024kling}, built on a DiT backbone, further advanced generative AI towards longer duration and greater visual quality.

Following the initial success of Generation 1 models in producing basic text–video consistency and plausible motion dynamics, some research shifted focus towards long and efficient video generation, sometimes at the expense of faithfulness. This led to the development of streaming video generation based on an autoregressive architecture. Specifically, LTX‑Video~\cite{hacohen2024ltxvideo} achieves semi‑real‑time generation, continuously producing video clips faster than they can be viewed, by leveraging a high‑compression Video‑VAE and a denoising transformer with full spatiotemporal attention. While notable for its generation speed, its limited fidelity of generated videos still places it within the first generation.

However, Generation 1 models lack strong 3D dynamics, making it difficult to maintain physically consistent motion, realistic object interactions, and task-oriented planning ability. 
As a result, generated videos may contain motion distortions, spatial misalignments, or other unrealistic artifacts that restrict their utility in more advanced interactive or long‑horizon planning tasks. These limitations motivate the developments in Generation 2 foundation models, which aim to enhance visual faithfulness, controllable interaction, and planning ability for more complex world‑modeling scenarios.

\subsubsection{Spatial World Model}
\label{sec:method_gen1_condition}


Although Any2Caption~\cite{wu2025any2caption} introduces the novel paradigm of “any-condition-to-caption” for video generation, the direct encoding of control signals and their integration with the video generation model remain essential. In particular, adopting a conditioned world model approach, where control signals are explicitly represented and fused within the generation process, provides a more intuitive and effective means of conditional guidance.
In this section, we introduce spatial world models with input $I = \{T, O, X\}$, which, although exhibiting some interaction capabilities, still suffer from limited video quality and controllability. Representative methods are summarized in Table~\ref{tab:General-methods} and Table~\ref{tab:app-methods}.

In general scenarios, geometry conditioned world models\cite{zhang2024moonshot,guo2024sparsectrl, zhang2023controlvideo, wang2023videocomposer,khachatryan2023text2video, peng2024conditionvideo, chen2023controlavideo} leverage conditional inputs that are semantically or structurally aligned with background content. Typical conditions include sketches, canny edges, depth maps, motion vectors, and human poses, extending the controllability paradigm established by ControlNet\cite{zhang2023controlnet} for image generation to the video domain. Representative works~\cite{guo2024sparsectrl, wang2023videocomposer, khachatryan2023text2video, liang2024diffusion4d, voleti2024sv3d, chen2024v3d} support text‑to‑video generation with geometry priors and further extend to other tasks such as animation~\cite{guo2024sparsectrl}, interpolation~\cite{wang2023videocomposer, khachatryan2023text2video} and 3D construction~\cite{voleti2024sv3d, chen2024v3d}. Notably, SparseCtrl~\cite{guo2024sparsectrl} introduces an image reference as geometry guidance for I2V generation, and due to its flexibility, it has been widely adopted as a baseline for controllable I2V/T2V generation under sparse conditioning.

3D prior world models~\cite{liang2024diffusion4d, voleti2024sv3d, chen2024v3d, park2025zero4d, yuan20244dynamic, zuo2024videomv} utilize 3D point cloud, explicit 3D modeling, or multi‑view imagery to improve spatial consistency in generated videos. While such priors are inherently tied to scene content and thus less adaptable across domains, their strong geometric grounding makes them more effective for producing 3D consistent videos and supporting 3D dynamics. Representative works include Diffusion4D~\cite{liang2024diffusion4d}, which incorporates explicit 3D point‑cloud priors within a video diffusion framework to achieve spatio‑temporal consistency and high‑fidelity 4D reconstruction, and Zero4D~\cite{park2025zero4d}, which employs explicit reconstructed 3D geometry from a single video in a training‑free manner to guide off‑the‑shelf video diffusion models for rapid 4D generation.

Physics prior world models incorporate physical signals or principles into video generation, enabling models to capture and reproduce realistic physical laws. For example, PhysGen~\cite{liu2024physgen} models forces, torques, and object interactions to simulate classical mechanics phenomena such as collisions governed by Newtonian dynamics. This approach allows generated videos to exhibit motion that is more consistent with real‑world physics, thereby enhancing their plausibility and potential for downstream simulation tasks.

Due to the characteristics of the tasks and application scenarios, early autonomous driving research featured several spatial conditioned world models compared with robotics and gaming. For example, Delphi~\cite{ma2024delphi} and Panacea~\cite{wen2024panacea} employed only scene layouts as spatial conditions to synthesize realistic driving videos, whereas MagicDrive~\cite{gao2023magicdrive} combined layouts, bounding boxes, and fixed camera‑pose parameters as a hybrid spatial condition.

\pengfeicom{Additionally, models~\cite{peng2024conditionvideo, zhang2023controlvideo, khachatryan2023text2video} condition on human pose are also categorize in spatial condition world model. The human pose we mention refers to the skeletal structure of a person in image space, represented as a sequence of keypoints that condition motion or structure within a fixed environment. Although human pose sequences have temporal form, they are spatially bound and depend on scene context, thus classified as spatial conditions rather than navigation conditions. In contrast, navigation conditions, such as trajectories or text commands are scene-independent and allow free navigation across different spatial contexts}

\subsubsection{Navigation World Model}
\label{sec:method_gen1_navigation}
Navigation world models inviting $N$ into the input set as $I = \{T, O, Au, N\}$. In general scenes, navigation world models are primarily developed for motion control tasks, including local subject motion~\cite{burgert2025gowiththeflow, geng2025motionprompting, namekata2024sgi2v, ma2024trailblazer, jain2024peekaboo, wang2024motionctrl, wu2024draganything, qiu2024freetraj, yang2024directavideo} and camera motion~\cite{bai2025recammaster, wang2025cinemaster, bahmani2025ac3d, bahmani2024vd3d, he2024cameractrl, van2024gcd, kuang2024cvd, bai2024syncammaster}. These methods often build upon pre-trained video foundation models~\cite{blattmann2023svd, wang2023modelscope, chen2023videocrafter1, yang2024cogvideox, bar2024lumiere, guo2023animatediff}, incorporating training-free control modules to enable interactive motion and camera control. However, due to current limitations in video length, generation quality, and the diversity of controllable entities, these approaches remain in the Generation 1 world models.
For local motion control tasks, DragAnything~\cite{wu2024draganything} offers intuitive point‑based dragging to finely adjust subject motion, enabling precise spatial manipulation in generated videos. Peekaboo~\cite{jain2024peekaboo} elevates this to semantic‑level control, where object‑centric motion can be directed in line with textual prompts. Pushing further, Trailblazer~\cite{ma2024trailblazer} incorporates trajectory‑driven navigation, supporting complex and continuous motion generation for both subjects and cameras, advancing towards more coherent and context‑aware video navigation.
In terms of camera control, existing methods~\cite{bahmani2024vd3d, bahmani2025ac3d, xu2024camco} inject camera motion trajectories, represented in Plücker coordinates, into pre-trained foundation models via ControlNet-based modules. Alternatively, methods such as GCD~\cite{van2024gcd} directly manipulate camera extrinsics, \ie, rotation and translation, as input to control camera viewpoints. 

Building on the general‑scene setting, navigational world models~\cite{gu2023seer, wu2024ivideogpt, zhou2024robodreamer, wu2024drivescape, ma2024delphi, wen2024panacea, gao2023magicdrive, jia2023adriveri, wang2024drivedreamer, wang2024drivewm, hu2022mile, yang2024genad, green2020mariovgg} have also been explored in more domain‑specific contexts such as robotics, autonomous driving, and gaming. For instance, in robotics, RoboDreamer~\cite{zhou2024robodreamer} incorporates additional modalities, such as sketches, to facilitate editable video generation, enabling interaction-level content manipulation. In autonomous driving, recent works, including ADriver‑I~\cite{jia2023adriveri}, DriveDreamer~\cite{wang2024drivedreamer}, Drive‑WM~\cite{wang2024drivewm}, MILE~\cite{hu2022mile}, and GenAD~\cite{yang2024genad}, further integrate high‑level scene semantics and temporal planning priors to enhance realism, consistency, and task‑oriented controllability.
In addition, in virtual gaming, playable video generation~\cite{menapace2021pvg} pioneers interactive, user‑controlled gameplay videos, enabling flexible manipulation of in‑game entities and camera perspectives. Moreover, MarioVGG~\cite{green2020mariovgg} focuses on faithfully recreating game scenes by learning domain‑specific visual dynamics, offering higher visual fidelity and consistency within structured game environments.
%


\subsection{Generation 2 - Interactiveness: Controllability and Interactive Dynamics}
\label{sec:methods_G2}

    Compared with Generation 1, Generation 2 world models achieve a marked leap in interactiveness, representing a decisive step towards dynamic and flexible interactive world modeling.
    Within this generation, interaction can be realized either through spatial conditions (\eg, sketches, depth, motion pose) or through navigation modes (\eg, trajectory, action, instruction).
    Spatial conditioned approaches~\cite{sketchvideo, das, bian2025gsdit, li2025pisa, xue2025phyt2v, wang2025wisa} excel in producing high‑quality videos with strong 3D dynamics, accurate condition–video consistency, and stable scene interactions, though their flexibility in real‑time control remains limited; nonetheless, these strengths firmly place them within the Generation 2 category.
    In addition, navigational world models~\cite{xiang2024pandora, aether, hong2024slowfast, hou2024camtrol, ren2025gen3c, he2025cameractrl2, fu20243dtrajmaster} offer greater flexibility and transferability, as they do not depend on pre‑existing contextual knowledge and can generalize across diverse entities and scenarios, making them inherently suitable for real‑time and general‑purpose applications.

    \begin{figure}[!t]
      \centering
      \includegraphics[width=\linewidth]{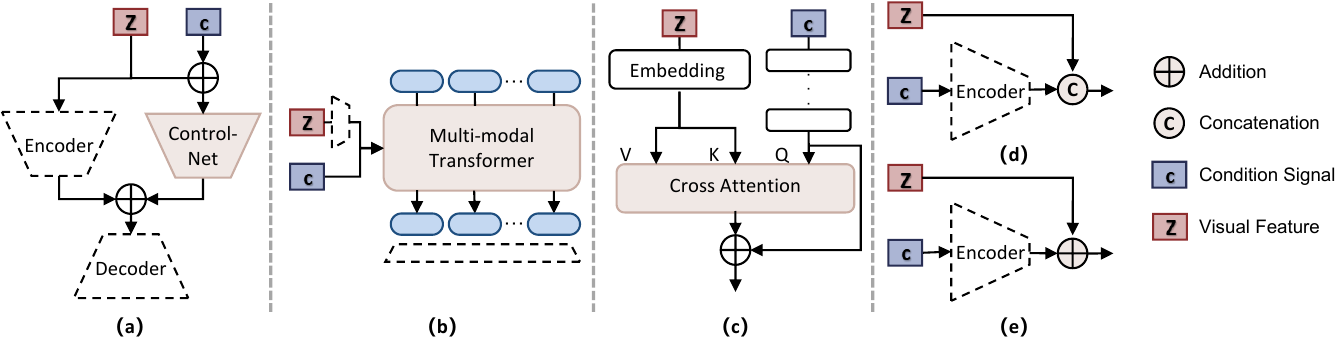}
      \caption{\textbf{Overview of Condition Injection Strategies.} This figure illustrates five representative strategies for condition injection. Subfigures (a) through (e) correspond to the ControlNet-based method, multi-modal Transformer, cross-attention, concatenation, and addition, respectively.}
      \label{fig:condition_inject}
    \end{figure}
    Starting from Generation 2, interactivity and controllability have become essential capabilities for video generation methods serving as world models. These requirements persist and evolve throughout Generation 3 and Generation 4, with increasingly higher demands placed on control granularity, control modes, modality diversity, real-time responsiveness, and control sensitivity. Given the advancement of controllable video generation models and the critical importance of integrating control signals into foundational video generation models in a principled manner, we systematically mentioned five major condition injection strategies: Condition Injection via ControlNet, Multi-modal Transformer, Cross-Attention, Concatenation, and Addition, with the fundamental architectures shown in Figure~\ref{fig:condition_inject}. 

    In this section, we review existing works in Generation 2, structured around world foundation models~\ref{sec:methods_G2_foundation} and four representative generative scenarios: general scenes (\ref{sec:methods_G2_general}), robotics (\ref{sec:methods_G2_rob}), autonomous driving (\ref{sec:methods_G2_ad}), and gaming (\ref{sec:methods_G2_gaming}). Within each scenario, we organize our discussion by tracing the progression from spatial condition-based methods to the more generalizable navigation-mode-based methods, further breaking down each category according to its defining characteristics and modeling approaches.

\subsubsection{Foundation Models with Dynamic-Aware World Knowledge}
\label{sec:methods_G2_foundation}

Building upon the first generation, Generation 2 foundation models make a decisive leap in interactiveness, representing a major step towards more dynamic and flexible world modeling. These advances stem from large‑scale pretraining, higher‑capacity architectures, and richer conditioning interfaces, enabling the models to adapt more effectively to downstream control mechanisms. As a result, spatial conditions and navigation‑mode controls can now be integrated with higher video fidelity, stronger spatiotemporal consistency, text-video consistency, greater control flexibility, and better 3D and semantic-level interaction, qualities that directly improve performance in downstream tasks across diverse scenarios.

Recent research builds on this momentum, consolidating video generation as the tool for controllable world models. \pengfeicom{These models~\cite{wan2025wan, peng2025opensora2, genmo2024mochi1,xu2024easyanimate, gao2025seedance, kong2024hunyuanvideo,sora,yang2024cogvideox, klingai2024kling, imaginegrok, wan22, openai2025sora2, luma2025ray3, team2025kling, bai2025semanticgen} commonly adopt DiT-based~\cite{peebles2023dit} or some of them leverage~\cite{lin2025apt2, yin2025causvid, huang2025selfforcing, ji2025memflow, nova25amazonaws, chen2025skyreels, teng2025magi} hybrid architectures that combine diffusion models with causal, frame-by-frame autoregressive mechanisms, achieving a fast, few-step distilled diffusion process to generate each frame. For instance, models such as MAGI‑1~\cite{teng2025magi} and CausVid~\cite{yin2025causvid} extend Diffusion Forcing~\cite{chen2024diffusionforcing} for causal video generation.} Self Forcing~\cite{huang2025selfforcing} follows this paradigm but removes exposure bias by training with full autoregressive rollout that matches the true inference distribution, enabling efficient video‑level supervision and semi-real‑time generation. And VMoBA~\cite{wu2025vmoba} introduces a mixture-of-block attention mechanism to enhance spatiotemporal representation learning in video diffusion models, achieving improved generation quality and efficiency.

In addition, the Cosmos series~\cite{agarwal2025cosmos, alhaija2025cosmostransfer1} illustrates how a powerful foundation model, trained on diverse multimodal inputs, can serve as a digital twin of the physical world while maintaining tight integration with external control, memory, and planning modules. 
When applied to complex, interactive environments, Cosmos demonstrates consistent, goal‑aligned video generation across navigation and simulation tasks, underscoring how improved interactiveness and condition adaptability in Generation 2 lay the groundwork for more general‑purpose, controllable world models. In addition, Luma AI introduced the Luma - Dream Machine, in which Luma Ray2~\cite{luma24} is a next-generation video generative model that integrates ultra-realistic detail, coherent motion, and logical event sequencing, making it particularly powerful in building world models.

\subsubsection{Video Generation as World Model in General Scene}
\label{sec:methods_G2_general}
In this section, we introduce methods that transition from video generation to world modeling in general-purpose scenarios, with representative methods summarized in Table~\ref{tab:General-methods}. The term general follows its definition in the broader video generation, referring to settings that are not tailored to any specific downstream application. 
Our discussion focuses on two main groups of methods:
(1) Approaches with spatial condition~\cite{sketchvideo, das, bian2025gsdit, li2025pisa, xue2025phyt2v, wang2025wisa, wang2025multishotmaster, li2025vfxmaster, huang2025unityvideo}, which incorporate various priors such as Geometry Conditions, 3D Priors, and Physical Priors. (2) Approaches with navigation mode~\cite{fu20243dtrajmaster, he2025cameractrl2, ren2025gen3c, hou2024camtrol, aether, hong2024slowfast, xiang2024pandora, tong20254vwm}, which condition generation on forms of controllable input, including action, trajectory, camera motion, and text instruction.

\noindent \textit{\textbf{Geometry Conditioned World Model.}}
Geometry conditioned world models utilize conditions that are semantically or structurally aligned with the background content. These conditions include, but are not limited to, sketches, canny edges, depth maps, motion vectors, and human poses, inheriting the controllability paradigm introduced by ControlNet~\cite{zhang2023controlnet} for image generation. Compared to Generation 1 approaches, Generation 2 models achieve much finer-grained interaction and better temporal fidelity through sparse, yet semantically rich, conditioning signals. 

For example, SketchVideo~\cite{sketchvideo} enables sketch-based video generation and editing by injecting sparse geometric cues, \ie, keyframe sketches, into a pretrained DiT model, maintaining temporal consistency through inter-frame attention. Compared to Generation 1, it demonstrates significantly improved spatial controllability and interactive editing with minimal user input.

\noindent \textit{\textbf{3D Prior World Model.}}
3D prior world models typically aim at 3D or 4D scene reconstruction, leveraging 3D priors to enhance the performance of video generation tasks in construction-oriented settings. In addition to this construction perspective, some approaches~\cite{bian2025gsdit, das} introduce 3D information specifically to benefit video generation itself. For example, DaS~\cite{das} leverages 3D point clouds to generate 3D tracking videos of the subject, which are then used as prompts to guide high-quality video synthesis.

\noindent \textit{\textbf{Physical Prior World Model.}}
Physical prior world models~\cite{li2025pisa, wang2025wisa, xue2025phyt2v, zhao2025synthetic, ji2025physmaster} aim to enhance the physical plausibility of video generation by embedding explicit physical laws and principles into the modeling process. For instance, WISA~\cite{wang2025wisa} incorporates physical formulas and principles directly as embeddings. PhyT2V~\cite{xue2025phyt2v} leverages the physical reasoning capabilities of large language models (LLMs) to guide and improve the physical consistency of video generation models.

\noindent \textit{\textbf{Camera Motion Navigation World Model.}}
Camera Motion Navigation World Models~\cite{fu20243dtrajmaster, he2025cameractrl2, ren2025gen3c, hou2024camtrol, luo2025camclonemaster} also rely on trajectories for navigation. Compared to general trajectory navigation, the input trajectories here are derived from camera motion, in the form of explicit camera moving trajectories or implicit reference videos. Moreover, camera movement naturally simulates first-person perspectives of agents or humans, this has evolved into a distinct and rapidly developing subtask. Therefore, we discuss it separately in this section. For example, CameraCtrl II~\cite{he2025cameractrl2} introduces an efficient diffusion framework that supports dynamic video generation guided by explicit camera trajectories, enabling smooth scene exploration through autoregressive clip extension and lightweight camera injection. In contrast, GEN3C\cite{ren2025gen3c} further enhances 3D spatial consistency by constructing a global 3D cache from depth predictions, allowing precise camera control and structurally faithful rendering across complex motion paths, advancing the realism and controllability of camera-based navigation models.

\noindent \textit{\textbf{Instruction Navigation World Model.}}
Instruction Navigation World Models~\cite{aether, xiang2024pandora, hong2024slowfast, li2024zerohsi, wei2025univideo} utilize text-based instructions, but unlike conventional text prompts that describe background elements, such as scene type, weather, or task themes, these instructions instead guide the dynamic aspects of video generation. Specifically, they control motion-related factors of either a single or multiple subjects in the video, or the perspective dynamics of a first-person agent, which can be a human, robot, or camera.
For instance, models like Pandora~\cite{xiang2024pandora} and SlowFast-Gen~\cite{hong2024slowfast} can execute directional commands such as “turn left” or “turn right.” In addition, SlowFast-Gen~\cite{hong2024slowfast} supports perspective-based actions such as zooming in and out from a navigational viewpoint.

In addition, AETHER~\cite{aether} unifies action-conditioned video prediction and visual planning within a single model from the agent’s first-person view. Remarkably, it also demonstrates strong zero-shot generalization capabilities.

\subsubsection{Video Generation as World Model in Robotics}
\label{sec:methods_G2_rob}

\definecolor{crText}{RGB}{243, 240, 161}    
\definecolor{crImage}{RGB}{247, 173, 190}   
\definecolor{crVideo}{RGB}{191, 155, 203}   

\definecolor{crBlur}{RGB}{240, 179, 210} 
\definecolor{crEdge}{RGB}{239, 208, 151}    
\definecolor{crDepth}{RGB}{162, 180, 156}  
\definecolor{crSegmentation}{RGB}{255, 200, 196} 
\definecolor{crSketch}{RGB}{197, 225, 222} 

\definecolor{crNone}{RGB}{238, 237, 234}    
\definecolor{crAction}{RGB}{240, 212, 209}  
\definecolor{crGoal}{RGB}{202, 210, 219}  
\definecolor{crCameraM}{RGB}{197, 225, 222} 

\definecolor{crTraj}{RGB}{250, 234, 219}
\definecolor{crVolume}{RGB}{160, 215, 209}  
\definecolor{crSmMap}{RGB}{219, 210, 201}   

\begin{table*}[!t]
    \setlength\extrarowheight{1pt}
    \centering
    \caption{\textbf{Summary and comparison of works for video generation as world model in robotics, autonomous driving, and gaming}. In the table, BBox refers to Bounding Box.
    }
    \label{tab:app-methods}
    {\fontsize{5pt}{7pt}\selectfont
    \begin{tabularx}{\linewidth}{
        p{0.2cm}p{3cm}p{3.5cm}p{3cm}p{2.3cm}X}
    \toprule
         
       & \bf{Method}  
       & \bf{Condition}
       & \bf{Navigation Mode}
       & \bf{Output Modality}    
       & \bf{Foundation Model} \\
    \midrule 
    \gray%
    \multicolumn{6}{c}{Generation 1 World Model} \\
    \midrule
    \multirow{4}{*}{\rotatebox{90}{Robotics}}
    & iVideoGPT~\cite{wu2024ivideogpt} 
    & None 
    & Action
    & Video, Action
    & -  \\

    & IRASim~\cite{zhu2024irasim}
    & None 
    & Trajectory
    & Video
    & Open-Sora~\cite{zheng2024opensora}  \\

    & Seer~\cite{gu2023seer}
    & None
    & Instruction
    & Video
    & - \\
    
    & RoboDreamer~\cite{zhou2024robodreamer}
    & Sketch
    & Instruction, Goal
    & Video
    & - \\

    \midrule
    \multirow{9}{*}{\rotatebox{90}{Autonomous Driving}}
        & MagicDrive~\cite{gao2023magicdrive}
        & Layout, BBox, Camera Position
        & None 
        & Video
        & -\\
        
        & DriveScape~\cite{wu2024drivescape}
        & Layout, BBox
        & None
        & Video
        & SVD~\cite{blattmann2023svd} \\

        & Delphi~\cite{ma2024delphi}
        & Layout
        & None
        & Video
        & - \\
        
        & Panacea~\cite{wen2024panacea}
        & Layout
        & None
        & Video
        & - \\

        & MILE~\cite{hu2022mile}
        & Layout
        & Action
        & Video, Action, Layout
        & -\\

        & Drive-WM~\cite{wang2024drivewm}
        & BBox
        & Action
        & Video, Action
        &-\\
        
        & DriveDreamer~\cite{wang2024drivedreamer}
        & BBox, Geometry
        & Action
        & Video, Action
        & - \\

        & ADriver-I~\cite{jia2023adriveri}
        & None
        & Instruction
        & Video
        & - \\

        & GenAD~\cite{yang2024genad}
        & None
        & Instruction, Trajactory
        & Video
        & -\\
    \midrule
        \multirow{2}{*}{\rotatebox{90}{Gaming}}
        & MarioVGG~\cite{green2020mariovgg} 
        & None  
        & Controller
        & Game Level
        & AVDC~\cite{ko2023avdc} \\

        & PVG~\cite{menapace2021pvg}
        & None
        & Controller
        & Video
        & -\\

        & Playable Environments~\cite{menapace2022playableenvironment}
        & None
        & Controller
        & Video
        & -\\


    \midrule
    \gray%
    \multicolumn{6}{c}{Generation 2 World Model} \\
    \midrule

    \multirow{19}{*}{\rotatebox{90}{Robotics}}
        & Cosmos-Transfer1~\cite{alhaija2025cosmostransfer1} 
        & Blur, Edge, Depth, Segmentation
        & None              
        & Video
        & Cosmos~\cite{agarwal2025cosmos} \\

        & PAD~\cite{guo2024pad} 
        & None
        & Action
        & Video, Action
        & - \\
        
        & WPE~\cite{quevedo2025wpe}
        & None
        & Action
        & Video
        & - \\

        & UVA~\cite{li2025uva}
        & None
        & Action
        & Video, Action
        & - \\
        
        & DWS~\cite{he2025dws}
        & None
        & Action
        & Video 
        & Open-Sora~\cite{zheng2024opensora}\\

        & 
        & None
        & Action
        & Video
        & iVdeoGPT~\cite{wu2024ivideogpt} \\
        
        & HMA~\cite{wang2025hma}
        & None
        & Action
        & Video, Action
        & SVD~\cite{blattmann2023svd} \\
        
        & Genie Envisioner~\cite{liao2025genieEnvisioner}
        & None
        & Action                  
        & Video
        & - \\
        
        & EVA~\cite{chi2024eva}
        & None
        & Instruction
        & Video
        & DynamiCrafter~\cite{xing2024dynamicrafter} \\

        & COMBO~\cite{zhang2024combo}
        & None
        & Instruction
        & Video
        & AVDC~\cite{ko2023avdc} \\

        & VideoAgent~\cite{fan2024videoagent}
        & None
        & Instruction
        & Video, Action
        & AVDC~\cite{ko2023avdc} \\

        & VidMan~\cite{wen2024vidman}
        & None
        & Instruction
        & Video, Action
        & Open-Sora~\cite{zheng2024opensora} \\
        
        & UniPi~\cite{du2023unipi}
        & None
        & Instruction
        & Video
        & Cliport~\cite{shridhar2022cliport}\\

        & VPDD~\cite{he2024vpdd}
        & None
        & Instruction
        & Video
        & -\\

        & AVDC~\cite{ko2023avdc}
        & Depth
        & Goal
        & Video, Action
        & - \\

        & Luo et.al~\cite{luo2024videotoaction}
        & None
        & Goal
        & Video
        & AVDC~\cite{ko2023avdc} \\

        & GR-1~\cite{wu2023gr1}
        & None
        & Action, Instruction
        & Video, Action
        & -  \\

        & RoboMaster~\cite{fu2025robomaster}
        & None
        & Action, Instruction          
        & Video
        & CogVideoX~\cite{yang2024cogvideox} \\

        & UniSim~\cite{yang2023unisim}
        & None
        & Action, Instruction, Camera
        & Video
        & -\\

        & NavigeteDiff~\cite{qin2025navigatediff}
        & None
        & Instruction, Goal
        & Video, Text
        & - \\
    \midrule

        \multirow{16}{*}{\rotatebox{90}{Autonomous Driving}}
        & Cosmos-Drive-Dreams~\cite{wu2024drivescape}
        & Layout
        & None
        & Video
        & Cosmos~\cite{agarwal2025cosmos} \\
        
        & UniScene~\cite{li2025uniscene}
        & Layout
        & None                
        & Video, LiDAR 
        & SVD~\cite{blattmann2023svd} \\

        & DreamForge~\cite{mei2024dreamforge}
        & Layout, BBox, Camera Position
        & None  
        & Video
        & - \\
        
        & DiVE~\cite{jiang2024dive}
        & Layout, Geometry
        & None
        & Video 
        & Open-Sora~\cite{zheng2024opensora} \\

        & DrivingDiffusion\cite{li2024drivingdiffusion} 
        & Layout, Optical
        & None
        & Video
        & - \\

        & DrivingGPT~\cite{chen2024drivinggpt}
        & None
        & Action
        & Video
        & -  \\

        & InfinityDrive~\cite{guo2024infinitydrive}
        & None
        & Action
        & Video
        & - \\
    
        & GAIA-1~\cite{hu2023gaia1}
        & None
        & Action
        & Video
        & - \\

        & Epona~\cite{zhang2025epona}
        & None
        & Trajectory 
        & Video, Trajectory 
        & - \\
        
        & GEM~\cite{hassan2025gem}
        & Geometry
        & Trajectory
        & Video
        & SVD~\cite{blattmann2023svd} \\

        & DrivingWorld~\cite{hu2024drivingworld}
        & None
        & Trajectory
        & Video
        & - \\
        
        & MagicDrive-V2~\cite{gao2024magicdrivedit}
        & Layout, BBox, Camera Position
        & Trajectory
        & Video
        & CogVideoX~\cite{yang2024cogvideox} \\

        & DriveDreamer-2~\cite{zhao2025drivedreamer2}
        & None
        & Instruction
        & Video
        & SVD~\cite{blattmann2023svd} \\

        & GAIA-2~\cite{russell2025gaia2}
        & Layout, Geometry
        & Action, Instruction
        & Video
        & - \\
       
        & MaskGWM~\cite{ni2025maskgwm}
        & None
        & Action, Trajectory, Instruction, Goal
        & Video
        & - \\
        
        & Vista~\cite{gao2024vista}
        & None
        & Action, Trajectory, Instruction, Goal
        & Video
        & SVD~\cite{blattmann2023svd} \\

    \midrule
    \multirow{14}{*}{\rotatebox{90}{Gaming}}
        & Hunyuan-GameCraft~\cite{li2025hunyuangamecraft}
        & None
        & Controller
        & Video
        & PlayGen~\cite{yang2024playgen} \\

        & MaaG~\cite{chen2025maag}
        & None
        & Controller
        & Game Level
        & PlayGen~\cite{yang2024playgen} \\

        & GameNGen~\cite{valevski2024gamengen} 
        & None
        & Controller
        & Video
        & - \\
        
        & W.H.A.M.~\cite{kanervisto2025wham}
        & None
        & Controller
        & Video
        & - \\
        
        & Genie~\cite{bruce2024genie}
        & Sketch
        & Controller
        & Video
        & - \\

        & PlayGen~\cite{yang2024playgen}
        & None
        & Controller
        & Video
        & - \\

        & GameGen-X~\cite{che2024gamegenx} 
        & None
        & Keyboard
        & Video
        & -\\

        & The Matrix~\cite{matrix} 
        & None
        & Keyboard
        & Video
        & - \\

        & MineWorld~\cite{guo2025mineworld} 
        & None
        & Keyboard, Mouse
        & Video
        & - \\

        & GameFactory~\cite{yu2025gamefactory}
        & None
        & Keyboard, Mouse
        & Video
        & - \\

        & WORLDMEM~\cite{xiao2025worldmem}
        & None
        & Keyboard, Mouse, Camera 
        & Video
        & - \\
        
        & DIAMOND~\cite{alonso2024diamond}
        & None
        & Keyboard, Mouse
        & Video, Action
        & EDM~\cite{edm22} \\

        & Genie-2~\cite{parker2024genie2}                         
        & None 
        & Keyboard, Mouse
        & Video
        & - \\

        & Oasis~\cite{decart2024oasis}                     
        & None
        & Keyboard, Mouse
        & Video
        & - \\

    \bottomrule
    \end{tabularx}
    }
\end{table*}

Due to the inherently interactive nature of robotics, it has naturally aligned with Generation 2 world models that utilize navigation modes from the very beginning of incorporating video generation models. Purely spatial conditioned approaches~\cite{alhaija2025cosmostransfer1} are rarely adopted in this domain, as they offer little practical value.

In robotics, three primary types of navigation modes are commonly used: action, text instruction, and goal.
The action mode typically refers to low-level physical signals such as force and torque applied by robotic arms.
The text instruction mode involves simple, short-horizon commands, \eg ``pick up the pen", while goal navigation encompasses more complex tasks that require long-term planning, such as ``clean up the desk." Goal navigation may also be defined by a target image representing the desired end state.

For clarity, we refer to short-horizon prompts as text instructions and use the term goal navigation for tasks that require planning over extended temporal horizons, including both textual goals and image-based goal points.

\noindent \textit{\textbf{Action Navigation World Model.}}
Action navigation world models~\cite{quevedo2025wpe, li2025uva, he2025dws, wang2025hma, guo2024pad} typically involve the prediction of actions as output, or alternatively, leverage video generation models to assist in forecasting the next action. For example, UVA~\cite{li2025uva} and PAD~\cite{guo2024pad} employ Transformer architectures to jointly encode video latent features and encoded actions into a unified representation, which is then decoded into both future video frames and an action policy.

\noindent \textit{\textbf{Instruction \& Goal Navigation World Model.}}
In the domain of text-guided robotics methods, world models~\cite{chi2024eva, zhang2024combo, fan2024videoagent, wen2024vidman, du2023unipi, he2024vpdd} at the action planning level have already begun to exhibit characteristics of basic planning capability in Generation 2. Since the introduction of AVDC~\cite{ko2023avdc}, there has been rapid progress in robotics world models capable of goal-directed planning based on image goals or textual instructions. For instance, COMBO~\cite{zhang2024combo} builds a compositional world model for multi-agent cooperation by factorizing joint actions and generating video predictions to simulate diverse outcomes. In contrast, UniPi~\cite{du2023unipi} treats decision-making as a text-conditioned video generation task, where a goal described in natural language is translated into future visual trajectories, from which control actions are extracted.

In contrast, methods such as Dreamitate~\cite{liang2024dreamitate} adopt a two-stage approach, which first uses a video generation model to produce high-quality future video sequences, which are subsequently utilized as input for action prediction. This separation of video synthesis and policy learning allows the system to benefit from strong visual foresight, enhancing decision-making performance in complex environments.

\noindent \textit{\textbf{Hybrid Navigation World Model.}}
Recently, hybrid navigation methods~\cite{yang2023unisim, wu2023gr1, qin2025navigatediff} have emerged, enabling models to handle both simple text instructions and image-goal navigation simultaneously. Beyond the text modality, models like UniSim~\cite{yang2023unisim} and GR-1~\cite{wu2023gr1} support multi-modal navigation using both text and action inputs to guide video synthesis. These multi-control setups raise important questions about balancing robustness and controllability across modalities, as well as potential conflicts between control signals. Notably, UniSim~\cite{yang2023unisim} supports semi-real-time interaction at the video level, aligning with our vision for advanced forms of Generation 2 world models.

\subsubsection{Video Generation as World Model in Autonomous Driving}
\label{sec:methods_G2_ad}

Autonomous driving represents another critical application domain for world models. Existing world models in this context are primarily employed to synthesize large-scale, multi-scene datasets or to perform visual and policy prediction tasks, serving as potential solutions to advance autonomous driving towards Level 4 autonomy.

In this section, we focus on two major research directions: layout-conditioned world models and those that incorporate multiple navigation modes, including text instructions, trajectory control, action-driven interaction, and hybrid navigation strategies. A detailed comparison of representative methods is provided in Table~\ref{tab:app-methods}.

\noindent \textit{\textbf{Layout Prior World Model.}}
Layout Prior World Models~\cite{ren2025cosmosdrivedream, li2025uniscene, mei2024dreamforge, jiang2024dive, li2024drivingdiffusion} in Autonomous Driving primarily refer to models where the input condition encodes the spatial distribution of vehicles and obstacles on the road. This typically includes HD maps, depth information, 3D bounding boxes, and the most common form, BEV (Bird’s-Eye View) layouts.

Representative works like UniScene~\cite{li2025uniscene}, which directly synthesizes both videos and LiDAR point clouds from layout-based inputs, facilitating large-scale autonomous driving dataset generation. Additionally, models like DreamForge~\cite{mei2024dreamforge} introduce camera poses to better align the layout with the generated video, while Driving Diffusion~\cite{li2024drivingdiffusion} leverages optical flow, effectively serving as an alternative form of layout conditioning.

However, because such layout information often relies on annotations available only in curated datasets, these approaches may lack practicality for real-time deployment. Accurate and timely collection of layout annotations in real-world applications remains challenging. Therefore, we consider layout-based methods to be an early-stage bridge towards building full-fledged world models in the autonomous driving domain.

\noindent \textit{\textbf{Instruction Navigation World Model.}}
In autonomous driving scenarios, text instruction world models~\cite{zhao2025drivedreamer2}  focus on ego-centric viewpoint changes, such as executing a right-turn command. However, existing models~\cite{zhao2025drivedreamer2} mainly target video generation conditioned on single-step instructions. Their performance in handling multi-step commands or long-horizon navigation plans remains largely underexplored, particularly in terms of stability and robustness.

\noindent \textit{\textbf{Trajectory Navigation World Model.}}
Trajectory Navigation World Models~\cite{zhang2025epona, hassan2025gem, hu2024drivingworld, gao2024magicdrivedit} navigate the video generation process by conditioning on either the ego-vehicle trajectory~\cite{hassan2025gem, hu2024drivingworld, gao2024magicdrivedit} or the trajectories of surrounding vehicles~\cite{zhu2025eotwm}. Most of these approaches~\cite{hu2024drivingworld, gao2024magicdrivedit} inject trajectory information via trajectory-encoded sequences embedded through Transformers. Others, like GEM~\cite{hassan2025gem}, utilize trainable LoRA modules~\cite{hu2022lora} to incorporate trajectory-based navigation in a more flexible and adaptive manner.

\noindent \textit{\textbf{Action Navigation World Model.}}
In Action Navigation World Models~\cite{chen2024drivinggpt, guo2024infinitydrive, hu2023gaia1}, the term action refers to vehicle control signals such as steering and velocity. DrivingGPT~\cite{chen2024drivinggpt} adopts a GPT-style autoregressive framework to achieve action-navigated video prediction. Other models~\cite{guo2024infinitydrive} start from an initial state derived from a combination of textual descriptions and reference images, and then generate future video frames conditioned on action signals. GAIA-1~\cite{hu2023gaia1} represents an interactive autonomous driving world model that casts world modeling as an unsupervised sequence modeling problem, mapping multimodal inputs into discrete tokens and predicting the next token in the sequence to simulate future scenarios.

\noindent \textit{\textbf{Hybrid Navigation World Model.}}
With the rapid advancement of autonomous driving, there is a growing trend towards the development of unified navigation world models in this domain. Recent works~\cite{russell2025gaia2, ni2025maskgwm, gao2024vista} have begun to support multiple conditioning signals and navigation modes simultaneously, \eg, \cite{russell2025gaia2} combining text instructions with action signals, text with trajectories, or \cite{ni2025maskgwm, gao2024vista} even integrating all four modalities: text instructions, trajectories, action commands, and goal point images. This convergence indicates that autonomous driving world models are gradually evolving towards planning capabilities, aligning with the characteristics of third-generation world models as defined in this paper.

However, despite these advances, real-time interaction remains largely absent from current autonomous driving world models. This limitation significantly hinders their practical application as decision-making or assistive modules in real-world autonomous vehicles.

\subsubsection{Video Generation as World Model in Gaming}
\label{sec:methods_G2_gaming}

Among the three primary application domains, even within general scene world modeling, the gaming domain has witnessed the fastest and most mature development of world models. This is largely driven by the inherent demand for real-time interaction in games, which has led most existing models in this area to support real-time feedback and responsiveness. In particular, open-world video generation in games closely aligns with our vision of third- and even fourth-generation world models: constructing a powerful representation model embedded with world knowledge, and then leveraging a video renderer to observe the world from various perspectives and initial conditions. In this setting, the open-world game environment effectively functions as an observable virtual world, one that is both dynamic and interactive in real time. Table~\ref{tab:app-methods} shows the details of representative works in gaming with their navigation modes and control level.

However, despite this progress, current game-oriented world models, such as those discussed in~\cite{yu2025position}, still face significant limitations in planning capabilities, viewpoint generalization, physical consistency, and semantic world representation. Therefore, we regard them as advanced second-generation world models, with ample room for further advancement.

\noindent \textit{\textbf{Controller Navigation World Model.}}
In the gaming domain, key control represents a crucial navigation mode for world models. This includes both inputs from game controllers~\cite{ bruce2024genie, kanervisto2025wham, valevski2024gamengen} and keyboard-based~\cite{yu2025gamefactory, alonso2024diamond, parker2024genie2, decart2024oasis} interactions. For instance, models~\cite{valevski2024gamengen, kanervisto2025wham} rely on controller inputs for gameplay interaction, while other models~\cite{yu2025gamefactory, alonso2024diamond, decart2024oasis} adopt a combination of keyboard and mouse inputs to enable frame-level control over game environments. Naturally, the complexity and granularity of supported actions vary across these models.

Notably, Google's Genie series~\cite{bruce2024genie, parker2024genie2}  marks a significant advancement in this area. Genie~\cite{bruce2024genie} is the first generative interactive environment trained in an unsupervised manner using unlabelled Internet videos. It can synthesize videos frame by frame from diverse inputs, including text-to-image prompts, hand-drawn sketches, text descriptions, or even real-world photos, all of which can be modulated via real-time action inputs. Its successor, Genie 2~\cite{parker2024genie2}, serves as a large-scale foundation world model. It demonstrates emergent capabilities at scale, such as object manipulation, complex character animation, physical reasoning, and even the prediction of other agents’ behavior. The Genie series is thus progressively evolving towards the vision of an interactive open-world simulation model that underpins next-generation world modeling.

\subsection{Generation 3 - Planning: Modeling the Future Evolution of Complex Systems}
\jiamingcom{In the third generation of world models~\cite{nwm, assran2025vjepa2, shin2025motionstream}, planning emerges as the core capability that begins to bridge the gap between environment-centered physical simulation and agent-centered reasoning.
This stage encompasses both objective physical planning, such as navigation and scene evolution based on global environmental dynamics, and the potential of subjective or agentic planning, where the model internally simulates trajectories, intentions, or counterfactual futures from a first-person perspective.
In this sense, planning represents the first step towards integrating a mental world model component within a primarily physical framework, enabling the model to reason not only about how the world evolves, but also about why such evolutions may occur under different beliefs or goals.}

\jiamingcom{From a theoretical standpoint, this generation naturally aligns with the MDP or POMDP duality:
during large-scale training, models approximate a fully observable physical world and learn an objective latent prior; during inference, they operate under partial observations, performing subjective reasoning over possible futures conditioned on actions, prompts, or multimodal inputs.
Thus, Generation 3 represents the transition point where objective world knowledge and subjective inference begin to co-exist within a unified planning process.}

\jiamingcom{Only a limited number of methods currently reach this level of capability.
The boundary between Generation 2 and Generation 3 is marked by the emergence of real-time inference ($\ge$ 24 FPS) and navigation-aware generation, where the system can adaptively simulate world trajectories based on external or internal control signals.
Models that demonstrate temporally consistent control and near-real-time rendering are regarded as exhibiting emergent potential towards this generation, even if fully consistent physics and complex multi-step planning have not yet been achieved.}

\jiamingcom{Representative examples include NWM~\cite{nwm}, which enables robots to imagine trajectories in unknown environments from a single input image, effectively performing self-navigated, belief-driven video prediction where the input image defines the initial world state.
Similarly, Meta’s V-JEPA 2~\cite{assran2025vjepa2} predicts the evolution of physical scenes and supports zero-shot robot planning in unseen environments, demonstrating the integration of predictive physical knowledge with adaptive task reasoning.
Genie 3~\cite{google2025genie3} further exemplifies this generation’s capabilities, achieving real-time (24 FPS, 720 p) interaction with minute-scale visual memory, allowing for consistent scene dynamics even as objects move out of view.
It also supports promptable world events, enabling users or agents to modify environmental conditions such as weather or layout mid-simulation, which shows an early form of interactive agent–environment co-evolution.}

\jiamingcom{Despite these advances, current models remain limited in scalability, action richness, and long-duration temporal consistency. Nevertheless, planning in Generation 3 signifies the critical turning point where world models evolve from purely physical simulators towards agentic and mental world modeling, unifying the objective and subjective dimensions of intelligent simulation.}

\subsection{Generation 4 - Stochasticity: Modeling Outlier and Low-Probability Events}

While the third generation of world models primarily focuses on faithfully simulating regular and rule-governed
aspects of the physical world, they tend to favor high-probability trajectories or events during planning. However,
the real world is inherently stochastic, characterized not only by deterministic rules but also by the occurrence of
rare, unpredictable, and sometimes disruptive events. The fourth generation of world models aims to incorporate
this essential dimension of realism by enabling the modeling and simulation of low-probability, outlier events.
These rare events, though statistically infrequent, often play a disproportionately significant role in the evolution
of complex systems. Examples include genetic mutations leading to pathological outcomes such as cancer, or sudden
accidents and coincidences in everyday life, such as traffic collisions. A mature world model should thus strike
a principled balance between simulating the most likely futures and accounting for the possibility of unexpected
deviations. This necessitates a more probabilistic, uncertainty-aware formulation of planning, one that acknowledges
the diversity of real-world dynamics beyond the average case.
By embedding stochasticity into world modeling, Generation 4 moves closer to emulating the richness and unpredictability of the physical world, laying the foundation for agents capable of robust decision-making under uncertainty.

\jiamingcom{Beyond visual realism, future physical world models should incorporate multimodal sensory channels (\eg. audio) to more faithfully emulate the perceptual diversity of real-world environments.
Audio signals convey fine-grained prompts such as collisions, weather dynamics, or human and animal behaviors, which are often crucial for robust situational understanding.
Integrating synchronized auditory modeling into world simulations can thus significantly enhance both the realism and the cognitive richness of the environment, providing agents with additional modalities to reason, predict, and act intelligently under uncertainty.
In this sense, the emergence of audio–visual world models~\cite{wan2025wan, runway2024gen3, Veo325, openai2025sora2, imaginegrok, pika23} will likely coincide with the development of higher-level agent intelligence, as richer sensory feedback enables more adaptive and context-aware interaction within stochastic environments.}

The fourth generation of world models consists of two progressive stages.
In the basic stage, the model is capable of learning the probability distributions of real-world events, including the spontaneous generation of everyday low-probability scenarios such as traffic accidents, rainy or snowy weather, and balanced stochastic outcomes like coin flips, newborn gender ratios, or vehicle turning directions at intersections. Built upon this probabilistic foundation, the model is further able to simulate extremely rare but high-impact events and generate coherent long-term evolutions conditioned on these events. Examples include financial crises, volcanic eruptions, genetic mutations, or asteroid collisions, events that, while statistically negligible, can fundamentally reshape future trajectories.

While these efforts largely operate within the realm of human-scale physics, \ie, mesoscopic spatiotemporal modeling, we envision the advanced stage of Generation 4 models to extend planning capabilities towards both macroscopic and microscopic scales.

At the macroscopic scale, a world model should be able to simulate and plan over long-term horizons, \eg, forecasting the development of a system over ten years given an initial state. Importantly, such planning is not expected to occur in real time; rather, the model must compress these ten years into manageable durations (\eg, one hour or even ten minutes), requiring multi-scale temporal modeling. Instead of predicting every single frame or timestep, the model should focus on summarizing salient events. While some existing works begin to address long-horizon forecasting and event-keyframe distillation, progress remains nascent.

On the microscopic scale, tasks such as modeling biological phenomena (\eg, involuntary eye microsaccades) remain beyond the reach of current video generation models due to limited temporal resolution and a lack of reasoning capabilities for such fine-grained fluctuations. Yet, this level of precision is crucial for improving both video generation quality and physical realism, and for faithfully simulating real-world subtleties. Most critically, this domain is still vastly underexplored in current literature, presenting a significant opportunity for future research.

\subsection{Insight: Beyond Generation}
Beyond Generation 4, we envision a more advanced form of world models that are capable of simulating everything occurring at every time and everywhere. Starting from a given initial state, repeated stochastic rollouts of the model would yield diverse plausible outcomes, resembling the conceptual structure of parallel universes. In this ultimate form, world models could provide downstream agents with virtually unlimited training trajectories and interaction scenarios, compensating for the lack of training data in downstream task domains.

\jiamingcom{In addition, beyond purely visual simulation, the recent emergence of multimodal generation models~\cite{Veo325, pika23, imaginegrok, sora, meta2025vibes, runway2024gen3, wan22, luma2025ray3} with both vision and audio modalities indicates a clear trajectory towards unified audiovisual world models, where sound, narration, and perception are jointly simulated with physics-based consistency.}

Moreover, the current paradigm of world modeling is inherently Earth-centric: both environments and agents are restricted to observations on the Earth, where “world” is implicitly defined as “Earth.” However, the physical laws governing environments beyond our planet may differ significantly, most notably in their underlying physics. If future world models can generalize to arbitrary physical laws through fine-tuning or in a zero-shot manner, this would unlock an entirely new set of downstream tasks, such as cosmic simulation or autonomous satellite testing. Such a capability could profoundly advance human understanding of the broader universe.

 \section{Future World Model: Everything, Everywhere, Anytime Simulation}
 \begin{figure}[!t]
      \centering
      \includegraphics[width=0.5\linewidth]{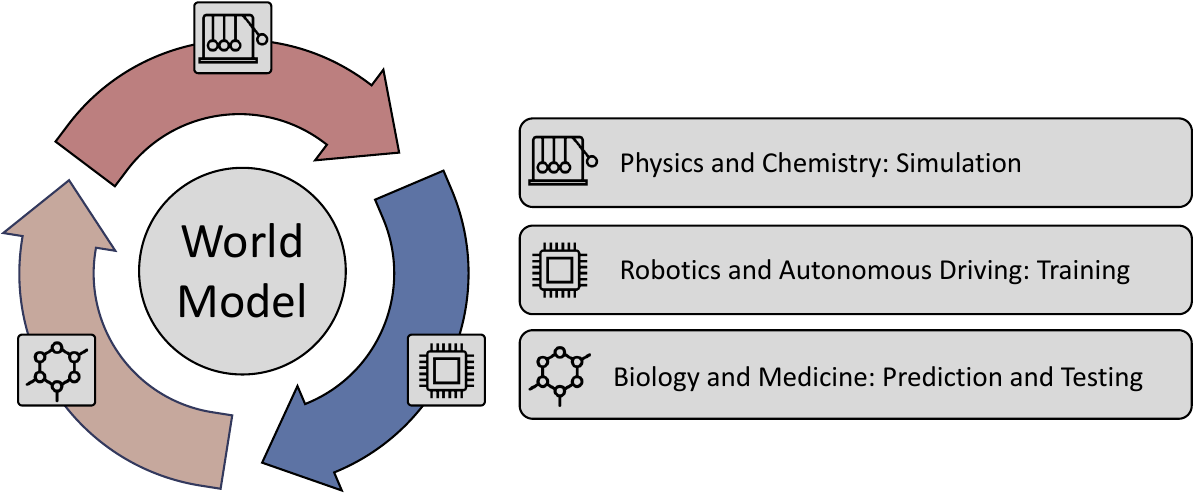}
      \caption{\textbf{Overview of Further Applications of World Models.} World models promise broad and long-term impact across diverse domains: simulating molecular structures and physical laws in physics and chemistry, generating synthetic training data and serving as virtual testbeds in robotics and autonomous driving, and enabling drug testing and protein structure prediction in biology and medicine.}
      \label{app}
    \end{figure}

\subsection{Two Complementary Development Directions}
\textbf{Precision Simulators. }
On the one hand, there is the pursuit of world models as accurate simulators. In this paradigm, the ultimate goal is to maximize fidelity to the real physical world, capturing its dynamics and stochasticity with unparalleled precision. Along this pathway, one might envision the eventual creation of a world model that can pass a “Turing Test for reality”: a system so accurate that its generated simulations are indistinguishable from actual observations of the physical world. Such models would serve as powerful scientific instruments, enabling researchers to validate hypotheses and test interventions in silico before deploying them in the real world.

\noindent
\jiamingcom{\textbf{World Models for Dicision and Control.}
In parallel with the vision and simulation centric trajectory of world modeling, another line of research has evolved within the reinforcement learning and robotics communities, where the focus lies on using learned world dynamics to support decision-making and control.
Rather than striving solely for pixel-level realism, these models emphasize predictive internal representations that enable agents to plan, imagine, and act within latent space.
Through learning compact transition dynamics, they allow an embodied system to anticipate outcomes of hypothetical actions, perform long-horizon reasoning, and optimize policies before real-world execution.
Notable examples such as PlaNet~\cite{hafner2019planet}, Dreamer Series~\cite{hafner2019dreamer, hafner2020dreamer2, hafner2023dreamerv3}, and related latent-dynamics models have demonstrated that predictive simulation and planning can effectively bridge perception and control, providing a complementary perspective to video-generation-based world modeling.
Together, these directions illustrate how world modeling can advance along both simulation fidelity and decision-oriented reasoning, converging towardsa more unified understanding of environment and agency.}

\noindent
\textbf{Generative Engines of World Knowledge. }On the other hand, we envision world models as engines of world knowledge and generative creativity. In this paradigm, the focus shifts from merely replicating a single reality to mastering world knowledge and enabling zero-shot generation of diverse possible world patterns. Such models, starting from a single initial state, could instantiate arbitrary virtual worlds, each governed by its own consistent set of physical or abstract laws. Importantly, multiple inferences from the same initial condition would yield divergent yet plausible outcomes, thereby generating parallel universes. In this sense, world models would empower individuals to `create and shape virtual worlds', not merely observing reality but actively creating new ones.

These two development directions, precision simulators versus creative generative engines, represent contrasting yet complementary visions. Together, they highlight the profound transformative potential of world models: both as tools for faithfully understanding our universe, and as platforms for exploring the infinite possibilities of imagined ones.

\subsection{Applications and Societal Impact}
Building upon the two directions of future world models realized based on the three core capabilities, direct models are poised to profound and potentially disruptive impact human modes of production and daily life, our ways of perceiving and understanding the world, the intellectual level of machine intelligence, and the methodologies employed by researchers across disciplines such as biology~\cite{han2025reusability, duan2025boosting, zhao2025protein}, physics~\cite{li2024electron, maurizi2025designing, nazari2025bioinspired}, astronomy~\cite{sanders2023biological, scott2023biomonitoring}, medicine~\cite{andani2025histopathology, ing2025integrating} and chemistry~\cite{li2025kolmogorov}.

Such models have the potential to address many of humanity’s challenges. On one hand, they tackle methodological and technological challenges. For example, in robotics~\cite{jung2024untethered, marcus2024ideal, seong2024multifunctional, dai2024autonomous, feng2024large, mao2024magnetic, huang2024programmable, mao2024multimodal, liu2024evolution, xia2024shaping}, an ideal world model could generate infinite real-world interaction data, resolving debates over whether developing better algorithms or collecting larger-scale data is more critical. In autonomous driving~\cite{abdel2024matched, liu2024curse}, it would allow us to directly simulate endless failure cases, greatly enhancing vehicle safety. On the other hand, world models could address challenges in application domains. For instance, they could predict wildlife~\cite{lee2024effects, goldberg2024widespread, buysse2024detection, watt2025parameters, baker2025dairy, burton2024mammal} habitats under varying conditions, monitor microbial growth states, and simulate atmospheric changes, thereby forecasting scenarios in which endangered species may face extinction, identifying protective measures that maximize survival and reproduction, predicting human behaviors that exacerbate global warming and extreme weather, and developing strategies to mitigate the intensifying effects of climate change~\cite{earth22025nvidia}. The applications extend even to physics, where such models could simulate multiple possible scenarios for cosmic formation or asteroid impacts on Earth.

\subsection{Conclusion and Outlook}
In conclusion, the evolution of world models promises to reshape the boundaries of human knowledge, creativity, and problem-solving. By integrating accurate simulation with generative and zero-shot capabilities, these models could serve as both a scientific laboratory and a virtual sandbox, enabling humanity to explore, understand, and intervene in complex systems at unprecedented scales. The pursuit of these dual capabilities represents one of the most ambitious frontiers in artificial intelligence and offers a vision of a future in which humans and machines co-create and navigate multiple possible worlds.

\section*{Acknowledgment}
We would like to thank Jiaming Song for the discussions and valuable feedback.
\newpage
\begin{appendices}




\end{appendices}

\bibliography{bibliography}

\end{document}